\pdfoutput=1

\documentclass[11pt]{article}

\usepackage[final]{acl}

\usepackage{times}
\usepackage{latexsym}

\usepackage[T1]{fontenc}

\usepackage[utf8]{inputenc}
\usepackage{multirow}
\usepackage{graphicx}
\usepackage{subcaption}
\usepackage{microtype}
\usepackage{hyperref}
\usepackage{inconsolata}

\usepackage{graphicx}
\usepackage{listings}
\usepackage{color}
\usepackage{xcolor}
\usepackage[normalem]{ulem}
\useunder{\uline}{\ul}{}
\usepackage{tcolorbox}
\tcbuselibrary{listings} 
\usepackage{booktabs}
\newtcolorbox[list inside=prompt,auto counter]{prompt}[1][]{
    colbacktitle=black!60,
    coltitle=white,
    fontupper=\footnotesize,
    boxsep=5pt,
    left=0pt,
    right=0pt,
    top=0pt,
    bottom=0pt,
    boxrule=1pt,
    #1,
}
\title{TROVE: A Challenge for Fine-Grained Text Provenance \\ via Source Sentence Tracing and Relationship Classification}

\author{
  Junnan Zhu$^{1}$\thanks{Corresponding Author.}\thanks{Equal Contribution.},
  Min Xiao$^{1,2}$\footnotemark[2],
  Yining Wang$^{4}$,
  Feifei Zhai$^{1,3}$,
  Yu Zhou$^{1,3}$,
  Chengqing Zong$^{1,2}$
  \\
  $^1$ State Key Laboratory of Multimodal Artificial Intelligence Systems, \\Institute of Automation, CAS, Beijing, China \\
  $^2$ School of Artificial Intelligence, University of Chinese Academy of Sciences, Beijing, China \\
  $^3$ Fanyu AI Laboratory, Zhongke Fanyu Technology Co., Ltd, Beijing, China \\
  $^4$ Unisound AI Technology Co.Ltd \\
  {\texttt \{junnan.zhu, yzhou, cqzong\}@nlpr.ia.ac.cn}\\
}

\begin{document}
\maketitle
\begin{abstract}
LLMs have achieved remarkable fluency and coherence in text generation, yet their widespread adoption has raised concerns about content reliability and accountability. In high-stakes domains, it is crucial to understand where and how the content is created. To address this, we introduce the Text pROVEnance (TROVE) challenge, designed to trace each sentence of a target text back to specific source sentences within potentially lengthy or multi-document inputs. Beyond identifying sources, TROVE annotates the fine-grained relationships (\emph{quotation}, \emph{compression}, \emph{inference}, and \emph{others}), providing a deep understanding of how each target sentence is formed.
To benchmark TROVE, we construct our dataset by leveraging three public datasets covering 11 diverse scenarios (e.g., QA and summarization) in English and Chinese, spanning source texts of varying lengths (0-5k, 5-10k, 10k+), emphasizing the multi-document and long-document settings essential for provenance. To ensure high-quality data, we employ a three-stage annotation process: sentence retrieval, GPT-4o provenance, and human provenance. We evaluate 11 LLMs under direct prompting and retrieval-augmented paradigms, revealing that retrieval is essential for robust performance, larger models perform better in complex relationship classification, and closed-source models often lead, yet open-source models show significant promise, particularly with retrieval augmentation. We make our dataset available here: \url{https://github.com/ZNLP/ZNLP-Dataset}.

\end{abstract}

\begin{figure*}[t]
\setlength{\abovecaptionskip}{0.12cm}    
\setlength{\belowcaptionskip}{-0.6cm}
  \centering 
  \includegraphics[width=\linewidth]{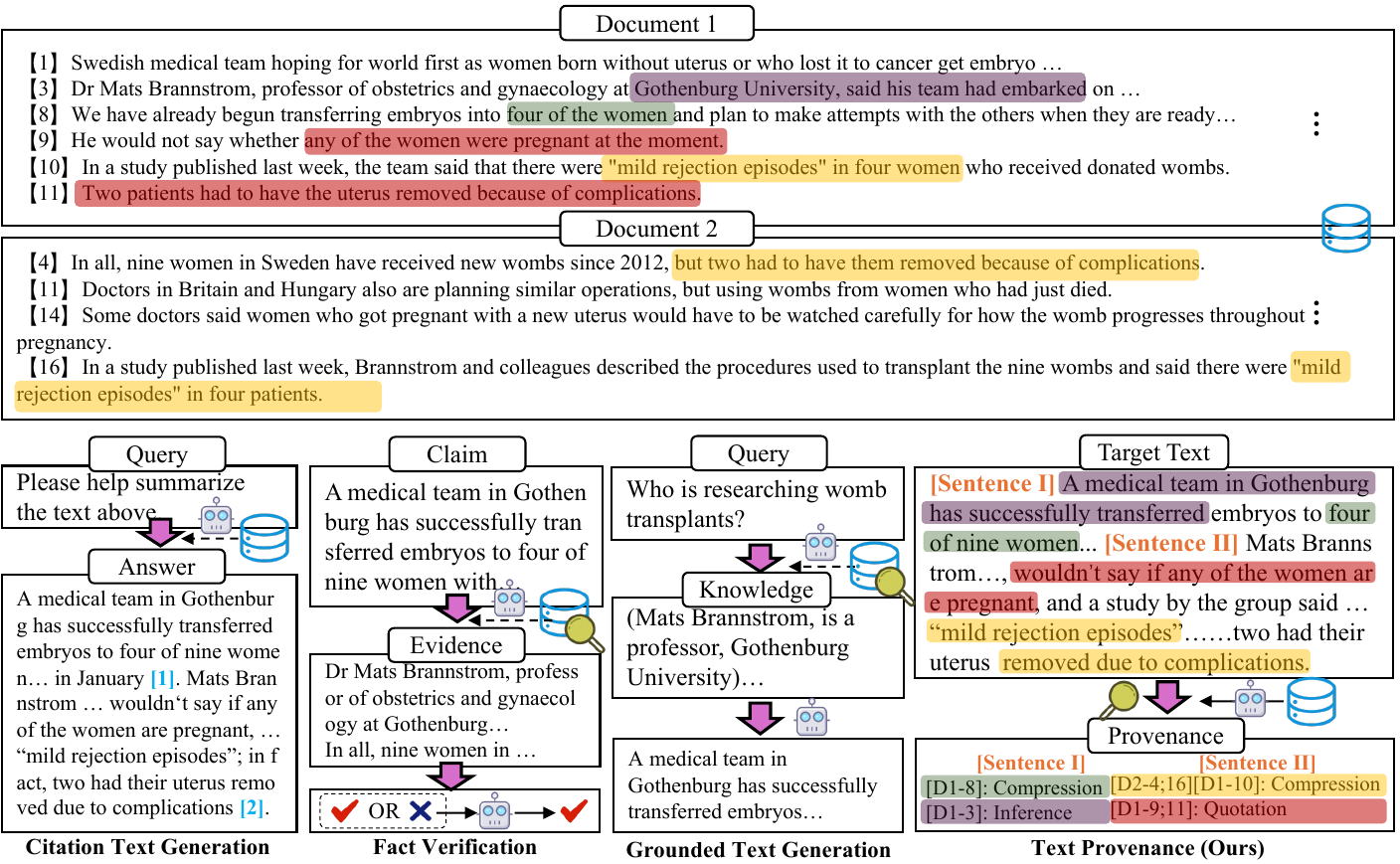}\\
  \caption{Overview of the difference between text provenance and related tasks. Solid arrows indicate required inputs and outputs, while dashed arrows represent optional ones. Compared with existing studies focusing on single-document or coarse-grained scenarios, our TROVE involves finer-grained provenance.\label{task}}
\end{figure*}

\section{Introduction}

Large language models (LLMs) have demonstrated great potential in natural language generation, producing highly coherent and fluent human-like text. However, their rapidly increasing prevalence raises significant concerns regarding content accountability and reliability. While considerable efforts have been made in citation~\cite{li-etal-2024-citation, huang-etal-2024-learning, cao-wang-2024-verifiable, aly-etal-2024-learning} and grounded generation~\cite{li-etal-2022-knowledge, brahman-etal-2022-grounded, slobodkin-etal-2024-attribute}, most existing studies focus on single-document-level source identification, leading to a significant gap in meeting the requirements of real-world scenarios. For instance, in domains like legal document drafting or medical reporting, it is crucial to identify where a sentence originates and understand how it has been generated from the sources. 

To bridge this gap, we introduce the challenge of text provenance (TROVE), which involves tracing a target text to the given source document(s) and establishing fine-grained relationships between the target and its source. TROVE is critical and challenging for large-scale information, as tracing sources becomes more complicated with longer or more numerous documents. To benchmark TROVE, we construct our dataset based on three public datasets: LongBench~\cite{bai-etal-2024-longbench}, LooGLE~\cite{li-etal-2022-knowledge}, and CRUD-RAG~\cite{crud2024}, considering the perspectives of multi-document, long-document, or their combination. 

Specifically, we first select examples with multi-sentence outputs to ensure sufficient sources and categorize data by original scenarios (question answering or summarization), languages (Chinese or English), input text length (0-5k, 5k-10k, or 10k+), and number of input documents (single-document or multi-document), aiming for a balanced distribution across each dimension. Next, we employ multiple information retrieval methods to recall the candidate source sentences for each sentence in the target text. Then, we use crafted prompts to guide GPT-4o in preliminary annotation, aiming to identify the sources of target sentences from retrieved candidate sentences and classify the target-source relationships into quotation (verbatim or partial copy), compression (summarization or paraphrase), inference (expansion, generalization, or specification), and others (e.g., negation). Each target sentence can be traced to multiple source sentences, and different relationships may apply simultaneously. For instance, a target sentence sourced from [a, b, c] might have [a, b] labeled as compression and [c] as quotation. Finally, we conduct a human review, requiring annotators to verify each annotation by considering both the source document(s) and the results produced by GPT-4o.

We perform a comprehensive evaluation of 11 models under two paradigms: \emph{direct prompting} and \emph{retrieval-augmented}, yielding valuable insights into the capabilities of current models in TROVE.

Our main contributions are as follows:
\begin{itemize}
	\item We introduce TROVE, a new challenge that traces each target sentence to its originating sources and classifies fine-grained target-source relationships beyond coarse-grained or single-document source identification.
	\item We present a carefully curated dataset covering multiple scenarios, languages, and source lengths. Our three-stage annotation produces high-quality, fine-grained provenance data.
	\item We systematically evaluate 11 LLMs (both closed-source and open-source) under multiple scenarios, revealing the necessity of retrieval augmentation, the advantages of larger models for relationship classification, and relationship classification remains challenging. 
\end{itemize}

\section{Related Work}
\textbf{Citation Text Generation}. Citation text generation focuses on producing text enriched with citations to enhance verifiability, making it widely applicable in academic writing~\cite{jurgens-etal-2018-measuring, xing-etal-2020-automatic, lauscher-etal-2022-multicite, mandal-etal-2024-contextualizing} and LLM-based chatbots~\cite{li-etal-2024-citation, huang-etal-2024-learning, cao-wang-2024-verifiable, aly-etal-2024-learning}. Existing approaches can be categorized into parametric and non-parametric methods. Parametric methods~\cite{mandal-etal-2024-contextualizing, gu-hahnloser-2024-controllable} rely on knowledge and patterns implicitly encoded within the model parameters to generate citation text. However, they face challenges in incorporating new citations or knowledge updates and are prone to hallucinations. Non-parametric methods~\cite{gao-etal-2023-enabling, huang-etal-2024-learning, li-etal-2024-citation} directly access external knowledge sources, such as citation databases, documents, or retrieval systems, to produce more reliable citation text. These approaches often leverage retrieval-augmented generation (RAG) techniques to integrate retrieved information with text generation. However, citations are typically generated in a post-hoc manner, which increases latency. Most existing methods focus on producing document-level, single-reference citations and emphasize the quality of the generated text. 

\textbf{Fact Verification}. Existing studies on fact verification~\cite{chen-etal-2024-complex, zheng-etal-2024-evidence, churina-etal-2024-improving} typically follow a two-stage approach: evidence retrieval and claim verification. Evidence retrieval aims to identify relevant passages or documents using information retrieval~\cite{chen2022gere, zheng-etal-2024-evidence} or neural ranking models~\cite{malviya-katsigiannis-2024-sk}. Claim verification aims to determine the authenticity of a claim by comparing it with the retrieved evidence, which has received more attention~\cite{zhong-etal-2020-reasoning, zou-etal-2023-decker}. 

\textbf{Grounded Text Generation}. Grounded text generation aims to produce text consistent with external sources of information, such as knowledge bases~\cite{li-etal-2022-knowledge, lu-etal-2022-controlling}, documents~\cite{slobodkin-etal-2024-attribute, hsu-etal-2024-calm}, or real-world facts~\cite{godbole-etal-2024-analysis, brahman-etal-2022-grounded}. This task ensures factual accuracy and coherence in generated content, as seen in applications like dialogue generation~\cite{li-etal-2022-knowledge, lu-etal-2022-controlling} and factual summarization~\cite{slobodkin-etal-2024-attribute, song-etal-2022-towards}. 

\textbf{Text Provenance: Unique Challenges and Contributions}. As shown in \autoref{task}, while citation text generation, fact verification, and grounded text generation all involve interactions between generated text and external sources, they each emphasize different aspects. Citation text generation focuses on incorporating references to support claims, fact verification aims to validate the truthfulness of statements, and grounded text generation ensures consistency with external information. In contrast, text provenance uniquely concentrates on identifying the specific source sentences for each target sentence and classifying the precise nature of their relationships, such as direct quotation, compression, inference, or negation. It requires retrieving relevant source sentences and performing a detailed semantic analysis to categorize the type of relationship, thereby providing a deeper understanding of how generated text originates from its sources. Consequently, text provenance extends beyond the capabilities of existing tasks by offering a more granular and relationship-focused approach to tracing the origins of generated content.


\section{Task Formulation}
Text Provenance aims to identify the source sentences for each target sentence in a generated text and classify the relationship between them using a given document collection. Specifically, given a target text $T=\{t_1, t_2, ..., t_n\}$, where each $t_i$ is a target sentence, and a document collection $D=\{d_1, d_2, ..., d_m\}$, where each documents $d_i$ contains a set of sentences $\{s_{i,1},s_{i,2},...,s_{i,k_i}\}$, the task is to determine, for each target sentence $t_i$, a set of source sentences $\{s_{i,j_1},s_{i,j_2},...,s_{i,j_k}\}$ from $D$ and classify their relationships.

Each target sentence $t_i$ may derive from multiple source sentences. The relationship between $t_i$ and each source sentences $s_{i,j}$ is categorized into one of the following types: \textbf{Quotation}, where $t_i$ is a verbatim or partial copy of $s_{i,j}$; \textbf{Compression}, where $t_i$ is a paraphrase or a summary derived from multiple source sentences, such as $s_{i,j_1}$ and $s_{i,j_2}$; \textbf{Inference}, where $t_i$ is logically inferred from one or more source sentences, such as $s_{i,j}$; \textbf{Others}, where $t_i$ does not fit the above categories, including cases like contradiction or negation.

For example, a target sentence $t_i$ may be derived from multiple source sentences, such as $s_{1,2}$, $s_{1,3}$, and $s_{2,1}$, where the relationship between $t_i$ and $s_{1,2}$, $s_{1,3}$ could be classified as compression, and the relationship between $t_i$ and $s_{2,1}$ could be classified as inference. This task thus requires a system to identify the appropriate source sentences and determine the precise relationship between each target sentence and its sources, which presents a complex challenge in understanding the fine-grained relationships between generated text and its origins.

\begin{figure*}[t]
    \centering
	\setlength{\abovecaptionskip}{0.12cm}    
	\setlength{\belowcaptionskip}{-0.6cm}
    \includegraphics[width=\linewidth,scale=1.00]{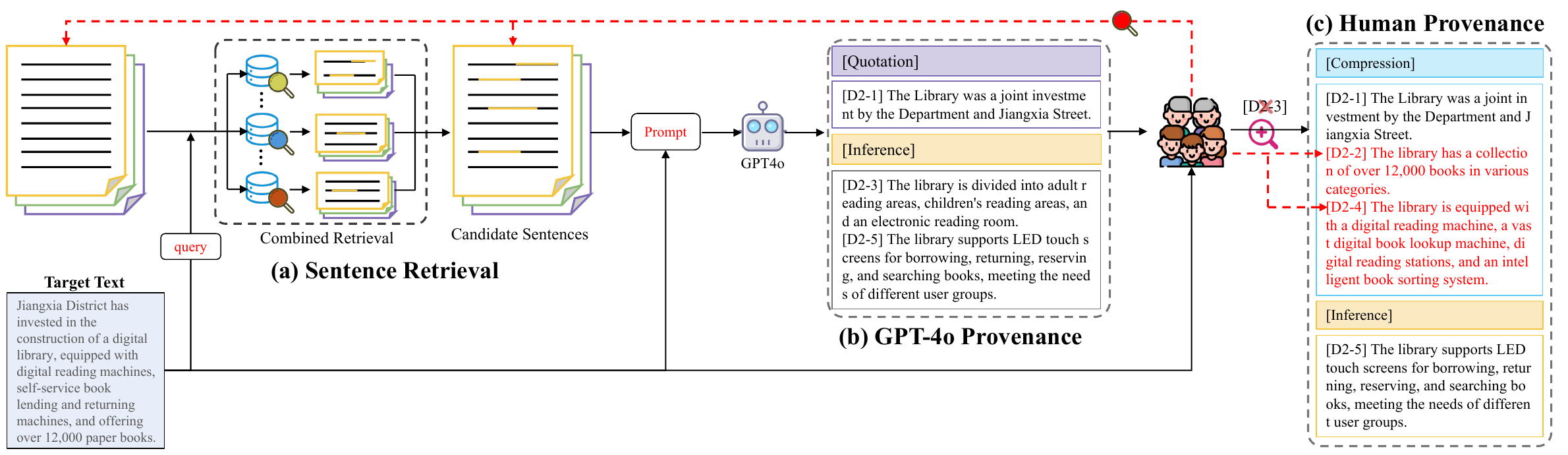}
    \caption{Overview of our data annotation. (a) Sentence Retrieval: selecting candidate provenance sentences using multiple retrievers; (b) GPT-4o Provenance: automatically annotating provenance relationships based on retrieved sentences; (c) Human Provenance: reviewing and refining GPT-4o’s annotations while independently checking source documents to identify missing provenance sentences. Di-j denotes the j-th sentence in the i-th document.}
    \label{fig:annotation}
\end{figure*}

\section{Datasets}
\subsection{Data Collection}
Text provenance becomes particularly crucial when dealing with large volumes of information, as tracing sources becomes more difficult with increasing document length and number. Therefore, we construct our text provenance dataset from the perspective of multi-document, long-document, or a combination of both, based on three public datasets: LongBench~\cite{bai-etal-2024-longbench}, LooGLE~\cite{li-etal-2022-knowledge}, and CRUD-RAG~\cite{crud2024}. 

LongBench focuses on long-context understanding and comprises 21 datasets in both English and Chinese across 6 task categories, covering key long-text application areas such as single-doc QA, multi-doc QA, and summarization. LooGLE is a comprehensive benchmark for evaluating long-context understanding in large language models. It features extremely long documents (post-2022) with over 24,000 tokens each and 6,000 questions across diverse domains, designed to assess short- and long-dependency tasks. CRUD-RAG is a comprehensive Chinese benchmark for evaluating Retrieval-Augmented Generation (RAG) systems. It categorizes RAG applications into four CRUD operations, i.e., Create, Read, Update, and Delete. It provides diverse evaluation tasks such as text continuation, question answering, hallucination modification, and multi-document summarization. 

From these datasets, we select examples where the output (or reference) contains multiple sentences, using these as the target text for our task. This approach ensures that the target text provides sufficient material for detailed source tracing, as each sentence may originate from different fragments of the input documents. We then treat each sample's original inputs as a unified document collection. Specifically, we sample from \textit{GovReport}, \textit{QMSum}, \textit{SAMSum}, \textit{VCSum}, and \textit{MultiNews} in LongBench and the long-dependency summarization (LongSum) task in LooGLE. Additionally, we include samples from \textit{EventSum}, \textit{QA1doc}, \textit{QA2doc}, and \textit{QA3doc} in CRUD-RAG. We categorize data by different tasks, languages (English and Chinese), input text length (0-5k, 5k-10k, and 10k+), and the number of input documents (single and multiple), trying to achieve a balanced distribution across these dimensions. Although we strive for a balanced distribution across these categories, some subsets inevitably remain underrepresented.

\begin{figure*}[t]
    \centering
	\setlength{\abovecaptionskip}{0.12cm}    
	\setlength{\belowcaptionskip}{-0.3cm}
    \includegraphics[width=\linewidth,scale=1.00]{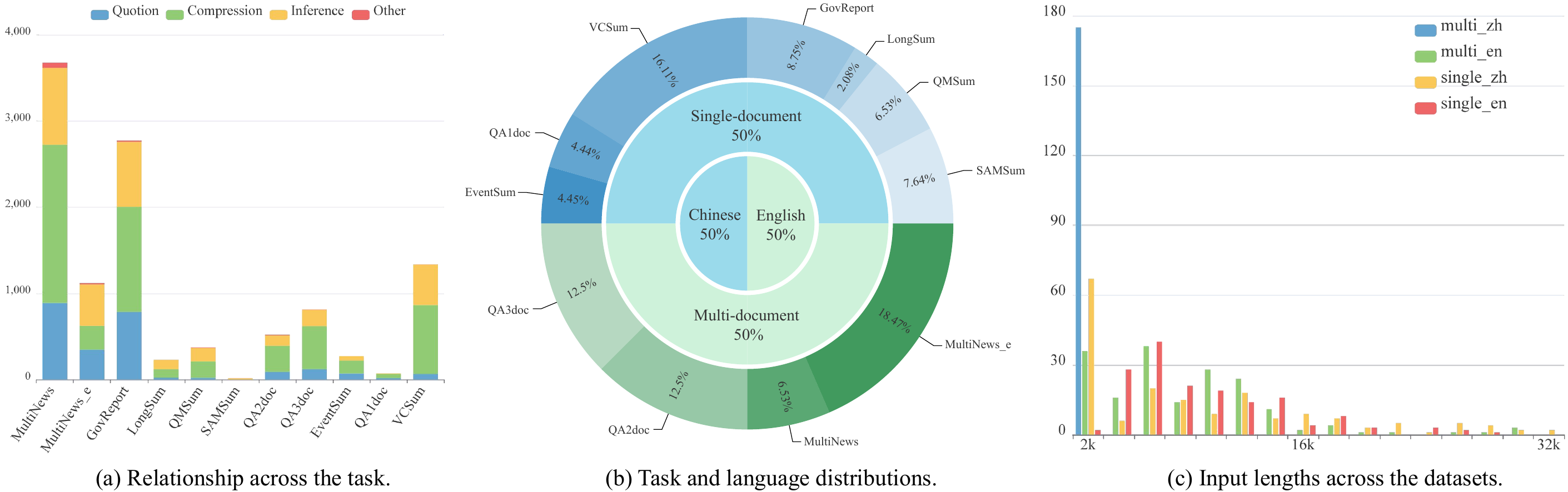}
    \caption{Dataset distribution.}
    \label{fig:distribution}
\end{figure*}

\begin{table*}[t]
\small
\centering
\setlength{\abovecaptionskip}{0.12cm}    
\setlength{\belowcaptionskip}{-0.5cm}
\begin{tabular}{cc|rrr|rr|rrrr}
\toprule
\multirow{2}{*}{\#Doc}  & \multicolumn{1}{l|}{\multirow{2}{*}{Lang}} & \multicolumn{3}{c|}{Source} & \multicolumn{2}{c|}{Target} & \multicolumn{4}{c}{Provenance}                                                                                                                                                                                                                        \\
                        & \multicolumn{1}{l|}{}                      & docs & sentences & tokens   & sentences      & tokens     & \begin{tabular}[c]{@{}c@{}}sentences\\ /example\end{tabular} & \begin{tabular}[c]{@{}c@{}}tokens\\ /example\end{tabular} & \begin{tabular}[c]{@{}c@{}}sentences\\ /sentence\end{tabular} & \begin{tabular}[c]{@{}c@{}}tokens\\ /sentence\end{tabular} \\ \midrule
\multirow{2}{*}{single} & zh                                         & 1.00 & 196.44    & 7,981.33 & 1.65           & 189.79     & 8.52                                                         & 509.21                                                    & 7.04                                                          & 421.56                                                     \\
                        & en                                         & 1.00 & 636.61    & 9,074.38 & 9.83           & 253.30     & 19.19                                                        & 620.33                                                    & 1.74                                                          & 52.46                                                      \\ \midrule
\multirow{2}{*}{multi}  & zh                                         & 2.51 & 20.95     & 903.48   & 2.44           & 146.02     & 6.78                                                         & 400.33                                                    & 2.98                                                          & 182.12                                                     \\
                        & en                                         & 3.77 & 327.93    & 7,222.53 & 12.13          & 320.47     & 23.10                                                        & 694.17                                                    & 1.97                                                          & 59.78                                                      \\ \bottomrule
\end{tabular}
\caption{Statistics of the dataset.}
\label{tab:statistics_dataset}
\end{table*}

\subsection{Data Annotation}

We employ GPT-4o to alleviate the manual annotation workload, as shown in \autoref{fig:annotation}. For each target sentence $t_i$, the annotation procedure consists of three steps: (a) sentence retrieval, (b) GPT-4o provenance, and (c) human provenance.

\textbf{Sentence Retrieval.} Due to the lengthy input text, GPT-4o may ignore key sentences when directly tracing provenance through long passages. To mitigate this, we first retrieve candidate provenance sentences using the target sentences as queries and then perform provenance based on these sentences. We have presented ablation studies to validate this approach.
Moreover, different retrievers capture diverse semantic features. To maximize the recall rate of candidate provenance sentences, we aggregate the results from $M$ distinct retrievers. Each retriever selects the top-$k$ most relevant sentences, denoted as $R_i(D, t_i)$, forming the following set of candidate provenance sentences:

\begin{small}
    \begin{equation}\label{equation:cand_sents}
    cands=\bigcup_i^M R_i(D,t_i)
    \end{equation}
\end{small}%
To reduce recall errors, only the union of sentences recalled by at least two retrievers is considered. We employ three retrievers: BM25~\citep{robertson2009probabilistic}, Dense~\citep{luan2021sparse}, and LCS~\citep{saadah2013information}. Each retriever selects the top-$k$ most relevant sentences, where $k=10$ .

\textbf{GPT-4o Provenance.} Based on the candidate provenance sentences, GPT-4o conducts fine-grained annotation and classifies the provenance relationship types, as depicted in \autoref{fig:annotation}(b). The detailed prompt is provided in the appendix.

\textbf{Human Provenance.} Annotators review GPT-4o’s results to verify the provenance sentences and their corresponding relationship types. Noting that GPT-4o can ignore critical details, the annotators examine the document collections to address any omissions. As illustrated in \autoref{fig:annotation}(c), the sentences ``D2-2'' and ``D2-4'', which contain ``12,000 books'' and ``digital reading machine'' respectively, exhibit a strong connection to the target sentence, yet GPT-4o fails to identify them. Therefore, the annotators will incorporate these two sentences into the final analysis. We invite 8 graduate students to spend about 510 hours annotating the provenance of 4,388 English and 811 Chinese sentences, costing an average of \$0.20 per sentence.

\subsection{Statistics of the Dataset}

\autoref{tab:statistics_dataset} provides detailed statistics of our dataset, including (1) Source: the average number of documents, sentences, and tokens per sample. (2) Target: the average number of target sentences and tokens per sample. (3) Provenance Results: the average number of provenance sentences and tokens per sample and the average number of provenance sentences and tokens per target sentence.

\autoref{fig:distribution} illustrates the dataset’s distribution across key characteristics: (a) relationship distributions across the task, (b) task and language distributions, and (c) source length distributions. These visualizations highlight the dataset’s high diversity.

\subsection{Consistency Analysis}
\begin{table}[t]
\small
\centering
\setlength{\abovecaptionskip}{0.12cm}    
\setlength{\belowcaptionskip}{-0.6cm}
\begin{tabular}{@{}cc|ccc@{}}
\toprule
\#Docs                  & Lang & Trace & Type  & GPT-4o \\ \midrule
\multirow{2}{*}{single} & zh   & .6696 & .5788 & .4391 \\
                        & en   & .6410 & .5336 & .5325 \\ \midrule
\multirow{2}{*}{multi}  & zh   & .7400 & .6187 & .5328 \\
                        & en   & .6004 & .4862 & .6997 \\ \bottomrule
\end{tabular}
\caption{Consistency of the annotation.}
\label{tab:consistency}
\end{table}

 To ensure dataset quality, 10\% of the examples are assigned to different annotators for consistency assessment. We evaluate annotator agreement from three perspectives: (1) tracing provenance sentences, (2) classification of relationship types, and (3) determination of necessary corrections to GPT-4o’s provenance sentences. To quantify agreement, we use Fleiss' Kappa~\citep{falotico2015fleiss} to measure the reliability across multiple annotators. The results, presented in \autoref{tab:consistency}, demonstrate that the annotation process is reliable.

 \begin{figure*}[t]
    \centering
	\setlength{\abovecaptionskip}{0.12cm}    
	\setlength{\belowcaptionskip}{-0.6cm}
    \includegraphics[width=\linewidth,scale=1.00]{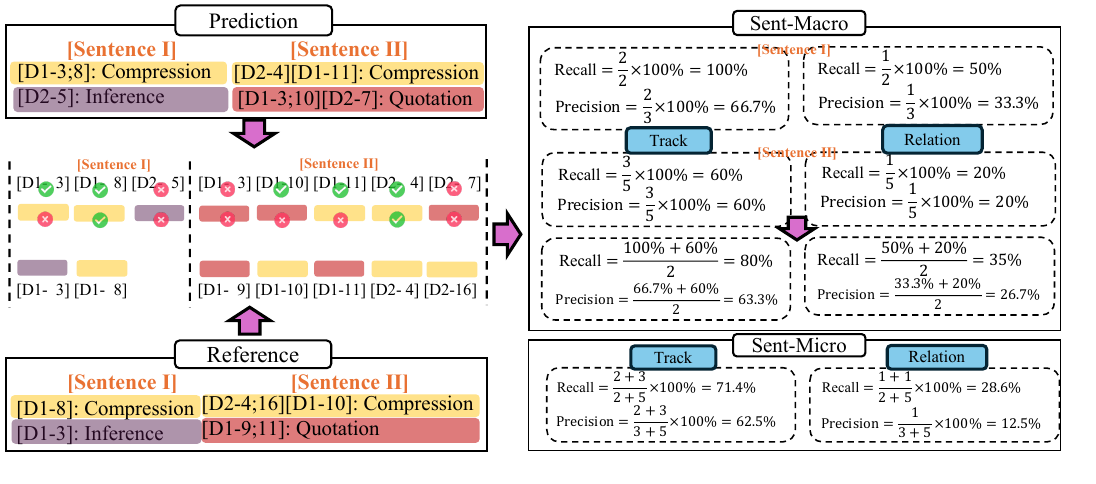}
    \caption{Overview of evaluation metrics for TROVE, including source tracing and relationship classification.}
    \label{fig:evaluation}
\end{figure*}
 
\section{Experiment}
\subsection{Experimental Setup}
We evaluate LLMs ranging from 6B to 671B parameters, including open-source and closed-source models. Open-source models include Qwen1.5-Instruct series~\citep{bai2023qwentechnicalreport}: Qwen1.5-Instruct-7B-chat, Qwen1.5-Instruct-14B-chat; Qwen2.5-Instruct series~\citep{qwen2025qwen25technicalreport}: Qwen2.5-Instruct-7B,  Qwen2.5-Instruct-14B; Llama-3-Instruct-8B~\citep{grattafiori2024llama3herdmodels}; ChatGLM2-6B~\citep{zeng2023glm130bopenbilingualpretrained}; Vicuna-7B-V1.5~\citep{vicuna}; DeepSeek-V3 (671B)~\cite{deepseekai2024deepseekv2}. Closed-source models include GPT-4o\footnote{\url{https://chat.openai.com/}}, Gemini-1.5-pro, Kimi\footnote{\url{https://kimi.moonshot.cn/}}.

The source length (0--32k) sometimes exceeds the context length supported by most LLMs. Therefore, we adopt a sliding-window approach for samples where the source text exceeds the model's maximum length limit, denoted as $M$. Specifically, the input text is split into chunks of $0-M, M-2\times M, 2\times M-3\times M$, etc. Each chunk is processed independently, and the final result is obtained by merging the predictions of all chunks.

During GPT-4o provenance, initial retrieval significantly enhances GPT’s recall rate. Thus, each model is evaluated under two approaches: (1) direct prompting tracing, where the model processes the input directly, and (2) retrieval-augmented tracing, where retrieval is performed first, followed by tracing based on the retrieved results.

\subsection{Provenance Automatic Evaluation}
We propose an evaluation method to assess model performance on this task, as shown in \autoref{fig:evaluation}.

First, to evaluate model accuracy in tracing target sentences and texts, we introduce macro-average and micro-average metrics at the sentence level. Macro-average metrics compute precision and recall for each sentence and average them across all target sentences in a sample. In contrast, micro-average metrics aggregate true predicted categories across all target sentences and calculate precision and recall based on global statistics.

In addition to evaluating source sentence tracing, we assess the model's ability to determine relationships between traced and target sentences. Our evaluation system includes 13 metrics for both source tracing and relationship classification: macro-average and micro-average precision and recall. Specifically, we compute \textbf{Macro-Track-P}, \textbf{Macro-Track-R}, \textbf{Micro-Track-P}, and \textbf{Micro-Track-R} for source tracing, as well as \textbf{Macro-Relation-P}, \textbf{Macro-Relation-R}, \textbf{Micro-Relation-P}, and \textbf{Micro-Relation-R} for relationship classification. To intuitively compare models, we calculate the F1 scores for Macro-Track-P, Macro-Relation-P, Micro-Track-P, and Micro-Relation-P, averaging them to derive the overall \textbf{F1-score}.


\subsection{Evaluation Results}

\definecolor{pink}{HTML}{FEF1E7}
\definecolor{green}{HTML}{EEF7E8}
\begin{table*}[t]
\centering
\setlength{\abovecaptionskip}{0.12cm}    
\setlength{\belowcaptionskip}{-0.6cm}
\resizebox{\textwidth}{!}{%
\begin{tabular}{@{}lcccccccccccccc@{}}
\toprule
\multirow{2}{*}{Model}                                                        & \multicolumn{1}{c|}{\multirow{2}{*}{Method}} & \multicolumn{6}{c|}{Macro}                                                                                                                                                                                                                     & \multicolumn{6}{c|}{Micro}                                                                                                                                                                                                           & \multirow{2}{*}{F1}              \\ \cmidrule(lr){3-14}
                                                                              & \multicolumn{1}{c|}{}                        & T-P                             & T-R                                       & T-F1                            & R-P                             & R-R                             & \multicolumn{1}{c|}{R-F1}                            & T-P                             & T-R                             & T-F1                            & R-P                             & R-R                             & \multicolumn{1}{c|}{R-F1}                            &                                  \\ \midrule
\multirow{4}{*}{Retrieval}                                                                     & \multicolumn{1}{c|}{LCS}                       & 19.71                            & 63.42                                      & 29.41                            & -                                & -                                & \multicolumn{1}{c|}{-}                                & 19.71                            & 61.28                            & 29.25                            & -                                & -                                & \multicolumn{1}{c|}{-}                                & 14.67                             \\ 
                                                                     & \multicolumn{1}{c|}{BM25}                       & 23.73                            & 78.66                                      & 35.70                            & -                                & -                                & \multicolumn{1}{c|}{-}                                & 23.73                            & 76.92                            & 35.56                            & -                                & -                                & \multicolumn{1}{c|}{-}                                & 17.81                            \\ 
                                                                     & \multicolumn{1}{c|}{Dense}                       & 17.85                            & 69.54                                      & 28.28                            & -                                & -                                & \multicolumn{1}{c|}{-}                                & 18.85                            & 67.41                            & 28.11                            & -                                & -                                & \multicolumn{1}{c|}{-}                                & 14.10                            \\ 
                                                                     & \multicolumn{1}{c|}{Union}                       & 33.89                            & 76.83                                      & 46.17                            & -                                & -                                & \multicolumn{1}{c|}{-}                                &  33.16                            & 74.99                             & 45.13                            & -                                & -                                & \multicolumn{1}{c|}{-}                                & 22.82                             \\ \midrule

\multicolumn{15}{c}{Open-Source}                                                                         \\ \midrule
\multirow{2}{*}{Vicuna-7b}                                                    & \multicolumn{1}{c|}{DP}                      & \phantom{0}6.80                  & 23.22                                      & 10.50                            & \phantom{0}2.37                  & \phantom{0}8.53                  & \multicolumn{1}{c|}{3.69}                            & \phantom{0}6.78                  & 23.10                            & 10.44                            & \phantom{0}2.38                  & \phantom{0}8.43                  & \multicolumn{1}{c|}{3.69}                            & 7.08                            \\
                                                                              & \multicolumn{1}{c|}{RA}                      & 27.14                            & 41.76                                      & 32.50                            & 9.79                             & 17.22                            & \multicolumn{1}{c|}{12.38}                            & 29.10                            & 40.78                            & 33.64                            & 10.39                            & 15.76                            & \multicolumn{1}{c|}{12.44}                            & 22.74                            \\ \midrule
\multirow{2}{*}{LLama3-8b}                                                    & \multicolumn{1}{c|}{DP}                      & \phantom{0}5.16                  & 16.63                                      & \phantom{0}6.97                  & \phantom{0}2.03                  & \phantom{0}7.21                  & \multicolumn{1}{c|}{\phantom{0}2.96}                  & \phantom{0}5.45                  & 15.50                            & \phantom{0}6.43                  & \phantom{0}1.84                  & \phantom{0}6.20                  & \multicolumn{1}{c|}{\phantom{0}2.49}                  & \phantom{0}4.71                  \\
                                                                              & \multicolumn{1}{c|}{RA}                      & 43.74                            & 38.19                                      & 40.61                            & 22.49                            & 19.51                            & \multicolumn{1}{c|}{20.82}                            & 49.81                            & 35.04                            & 41.07                            & 25.42                            & 18.40                            & \multicolumn{1}{c|}{21.33}                            & 30.96                            \\ \midrule
\multirow{2}{*}{Chatglm-6b}                                                   & \multicolumn{1}{c|}{DP}                      & \phantom{0}0.02                  & \phantom{0}0.04                            & \phantom{0}0.02                  & \phantom{0}0.01                  & \phantom{0}0.00                  & \multicolumn{1}{c|}{\phantom{0}0.01}                  & \phantom{0}0.04                  & \phantom{0}0.02                  & \phantom{0}0.03                  & \phantom{0}0.01                  & \phantom{0}0.01                  & \multicolumn{1}{c|}{\phantom{0}0.01}                  & 0.02                             \\
                                                                              & \multicolumn{1}{c|}{RA}                      & \phantom{0}3.68                  & \phantom{0}4.06                            & \phantom{0}3.84                  & \phantom{0}1.50                  & \phantom{0}1.76                  & \multicolumn{1}{c|}{\phantom{0}1.60}                  & 11.97                            & \phantom{0}3.98                  & \phantom{0}5.93                  & \phantom{0}4.85                  & \phantom{0}1.73                  & \multicolumn{1}{c|}{\phantom{0}2.53}                  & 3.47                             \\ \midrule
\multirow{2}{*}{Qwen1.5-7b}                                                   & \multicolumn{1}{c|}{DP}                      & \phantom{0}6.47                  & 41.00                                      & 11.18                            & \phantom{0}1.36                  & 11.71                  & \multicolumn{1}{c|}{\phantom{0}2.36}                            & \phantom{0}6.50                  & 40.82                            & 10.84                            & \phantom{0}1.27                  & 10.04                 & \multicolumn{1}{c|}{\phantom{0}2.20}                            & 6.65                            \\
                                                                              & \multicolumn{1}{c|}{RA}                      & 35.99                            & 53.25                                      & 42.26                            & 11.02                  & 19.08                            & \multicolumn{1}{c|}{13.80}                            & 34.83                             & 52.19                            & 41.00                            & 10.26                  & 16.48                            & \multicolumn{1}{c|}{12.48}                            & 27.39                            \\ \midrule
\multirow{2}{*}{Qwen2.5-7b}                                                   & \multicolumn{1}{c|}{DP}                      & \phantom{0}8.88                  & 49.56                                      & 14.77                            & \phantom{0}2.89                  & 15.49                            & \multicolumn{1}{c|}{4.81}                            & \phantom{0}7.67                  & 49.50                            & 12.94                            & \phantom{0}1.93                  & 13.56                            & \multicolumn{1}{c|}{3.29}                            & 8.95                            \\
                                                                              & \multicolumn{1}{c|}{RA}                      & 41.96                            & \textbf{\colorbox{green}{71.92}}           & 52.23                            & 14.73                            & 28.56                            & \multicolumn{1}{c|}{19.23}                   & 39.50                            & \textbf{\colorbox{green}{69.72}} & 49.65                            & 12.32                            & 24.55                            & \multicolumn{1}{c|}{16.21}                            & 34.33                            \\ \midrule
\multirow{2}{*}{Qwen1.5-14b}                                                  & \multicolumn{1}{c|}{DP}                      & 12.50                            & 38.97                                      & 16.84                            & \phantom{0}3.71                             & 13.35                            & \multicolumn{1}{c|}{\phantom{0}5.25}                            & 13.36                            & 37.51                            & 16.93                            & \phantom{0}3.65                  & 11.32                            & \multicolumn{1}{c|}{\phantom{0}4.74}                            & 10.94                            \\
                                                                              & \multicolumn{1}{c|}{RA}                      & 45.07                            & 53.68                                      & 48.33                            & 14.06                            & 20.96                            & \multicolumn{1}{c|}{16.72}                            & 47.99                            & 51.65                            & 49.20                            & 14.50                            & 17.90                            & \multicolumn{1}{c|}{15.93}                            & 32.54                            \\ \midrule
\multirow{2}{*}{Qwen2.5-14b}                                                  & \multicolumn{1}{c|}{DP}                      & 29.77                            & 52.25                                      & 36.12                            & 15.11                            & 27.37                            & \multicolumn{1}{c|}{18.69}                            & 29.50                            & 49.55                            & 33.01                            & 14.37                            & 24.74                            & \multicolumn{1}{c|}{16.24}                            & 26.02                            \\
                                                                              & \multicolumn{1}{c|}{RA}                      & \textbf{54.60}                   & 50.24                                      & \textbf{51.99}                   & \textbf{29.22}                   & 27.43                   & \multicolumn{1}{c|}{28.12}                            & \textbf{64.68}                   & 47.02                            & \textbf{54.23}                   & \textbf{33.49}                   & 24.80                   & \multicolumn{1}{c|}{\textbf{28.37}}                   & \textbf{40.68}                   \\ \midrule
\multirow{2}{*}{\begin{tabular}[c]{@{}l@{}}DeepSeek-V3\\ (671B)\end{tabular}} & \multicolumn{1}{c|}{DP}                      & 44.79                            & \multicolumn{1}{l}{\colorbox{pink}{56.56}} & 49.80                            & 21.88                            & 28.40                            & \multicolumn{1}{c|}{24.63}                            & 39.54                            & \colorbox{pink}{54.31}           & 44.95                            & 17.46                            & 25.61                            & \multicolumn{1}{c|}{20.41}                            & 34.95                            \\
                                                                              & \multicolumn{1}{c|}{RA}                      & 49.17                   & 55.85                                      & 51.94                   & 26.10                   & \textbf{31.20}                   & \multicolumn{1}{c|}{\textbf{28.24}}                            & 50.75                   & 53.76                            & 51.93                   & 26.17                   & \textbf{28.54}                   & \multicolumn{1}{c|}{27.19}                   & 39.83                   \\ \midrule
\multicolumn{15}{c}{Closed-Source}                                                                                                                                                                                                                                                                                                                                                                                                                                                                                                                                                                                                                      \\ \midrule
\multirow{2}{*}{GPT-4o}                                                       & \multicolumn{1}{c|}{DP}                      & \colorbox{pink}{59.31}           & 55.46                                      & \colorbox{pink}{57.18}           & \colorbox{pink}{36.55}           & \colorbox{pink}{33.98}           & \multicolumn{1}{c|}{\colorbox{pink}{35.15}}           & \colorbox{pink}{57.32}           & 52.43                            & \colorbox{pink}{54.55}           & \colorbox{pink}{34.39}           & \colorbox{pink}{31.61}           & \multicolumn{1}{c|}{\colorbox{pink}{32.81}}           & \colorbox{pink}{44.92}           \\
                                                                              & \multicolumn{1}{c|}{RA}                      & 73.14                            & 55.45                                      & 62.72                            & 42.68                            & 32.87                            & \multicolumn{1}{c|}{36.94}                            & \textbf{\colorbox{green}{74.81}} & 51.59                            & 60.84                            & 42.93                            & 30.25                            & \multicolumn{1}{c|}{35.38}                            & 48.97                            \\ \midrule
\multirow{2}{*}{Gemini-1.5-pro}                                               & \multicolumn{1}{c|}{DP}                      & 13.30                            & 13.00                                      & 13.10                            & \phantom{0}7.54                  & \phantom{0}7.44                  & \multicolumn{1}{c|}{\phantom{0}7.46}                            & 11.38                            & 11.71                            & 11.49                            & \phantom{0}6.29                  & \phantom{0}6.54                  & \multicolumn{1}{c|}{\phantom{0}6.40}                            & \phantom{0}9.61                            \\
                                                                              & \multicolumn{1}{c|}{RA}                      & \textbf{\colorbox{green}{74.13}} & 58.53                                      & \textbf{\colorbox{green}{64.94}} & \textbf{\colorbox{green}{45.02}} & 34.45                            & \multicolumn{1}{c|}{\textbf{\colorbox{green}{38.75}}} & 73.80                            & 54.62                            & \textbf{\colorbox{green}{62.43}} & \textbf{\colorbox{green}{46.00}} & \textbf{\colorbox{green}{33.52}} & \multicolumn{1}{c|}{\textbf{\colorbox{green}{38.58}}} & \textbf{\colorbox{green}{51.18}} \\ \midrule
\multirow{2}{*}{Kimi}                                                         & \multicolumn{1}{c|}{DP}                      & 39.75                            & 47.29                                      & 43.12                            & 20.38                            & 24.82                            & \multicolumn{1}{c|}{22.34}                            & 36.01                            & 44.65                            & 39.68                            & 17.52                            & 22.03                            & \multicolumn{1}{c|}{19.42}                            & 31.14                            \\
                                                                              & \multicolumn{1}{c|}{RA}                      & 60.27                            & \textbf{64.69}                             & 62.25                            & 32.44                            & \textbf{\colorbox{green}{36.05}} & \multicolumn{1}{c|}{34.07}                            & 57.25                            & \textbf{61.50}                   & 59.15                            & 30.32                            & 33.41                            & \multicolumn{1}{c|}{31.73}                            & 46.80                            \\ \bottomrule
\end{tabular}
}
\caption{Experiment results of LLMs. DP and RA denote direct prompting tracing and retrieval-augmented tracing. In both open-source and closed-source models, \colorbox{pink}{pink} denotes the best DP results, while \colorbox{green}{green} marks the best RA results. The \textbf{bold values} highlight the best results within open and closed-source models, respectively. Since the union retrieval method outperforms each single retrieval method, we use the union retrieval method in RA.}
\label{tab:experiment}
\end{table*}

\textbf{Impact of Retrieval-Augmented Tracing vs. Direct Prompting Tracing.} Across almost all models, retrieval-augmented tracing outperforms direct prompting in F1 scores, often by a large margin. For example, Qwen2.5-14B’s F1 jumps from 26.02 to 40.68 with retrieval, while ChatGLM-6B, which nearly fails in direct prompting tracing with an F1 of 0.02, improves to 3.47. Even closed-source models show the same trend, as Gemini-1.5-Pro significantly improves from an F1 of 9.61 to 51.18 with retrieval. It suggests that retrieval helps overcome context-length limits and brings in the relevant source text, making it much easier for models to match target sentences with their sources.

\textbf{Impact of Model Size}. As shown in \autoref{tab:experiment}, larger models generally achieve higher scores in source tracing and relationship classification. For example, Qwen2.5-14B (retrieval) outperforms its smaller counterparts in most metrics, such as Track-P and Relation-P. However, Qwen2.5-7B (retrieval) achieves the highest Track-R scores, indicating that smaller models can also perform well in specific aspects of source tracing even if they do not lead in the overall F1-score. While the trend favors larger models, specific architectures or training strategies allow smaller models to remain competitive in the provenance task. Notably, for relationship classification, the advantage of larger models is more consistent, suggesting that capturing complex relationships (such as paraphrasing, summarization, and logical inference) demands the enhanced representational capacity of increased parameterization.

\textbf{Precision–Recall Trade-offs Across Models.} We can find some interesting trade-offs when examining the precision and recall metrics for each model. Some models, like Qwen2.5-7B with retrieval, prioritize recall, identifying more traced sources with a recall of 71.92, but at the cost of lower precision at 41.96. Others, such as Qwen2.5-14B with retrieval, achieve a better balance, reaching a higher precision of 54.60 while maintaining a recall of 50.24. In real-world applications, a high-recall system may be preferable when capturing all possible source sentences, which is crucial, even if some false positives appear. On the other hand, a precision-focused system is better suited when avoiding false positives is a priority.

\textbf{Open-Source vs. Closed-Source.} Among open-source models, parameter sizes vary widely, from a few billion (e.g., 6B–14B) to the much larger Deepseek V3 with 671B parameters. Despite these differences, larger models generally perform better in direct prompting and retrieval-augmented settings, especially in relationship classification. Deepseek-V3 (DP) shows strong performance with an F1 score of 34.95, outperforming many smaller models. However, when retrieval is applied, models like Qwen2.5-14B begin to reduce the gap with leading closed-source systems. For closed-source models, Gemini-1.5-Pro (RA) and GPT-4o (RA) achieve the highest F1 scores at 51.18 and 48.97, performing well in both source tracing and relationship classification. However, Gemini-1.5-Pro struggles with direct prompting, with an F1 score of only 9.61, highlighting the importance of retrieval. While closed-source models still lead overall, their advantage is reduced significantly when open-source LLMs use strong retrieval methods.


\begin{figure*}[t]
    \centering
	\setlength{\abovecaptionskip}{0.12cm}    
	\setlength{\belowcaptionskip}{-0.4cm}
	\includegraphics[width=0.985\linewidth]{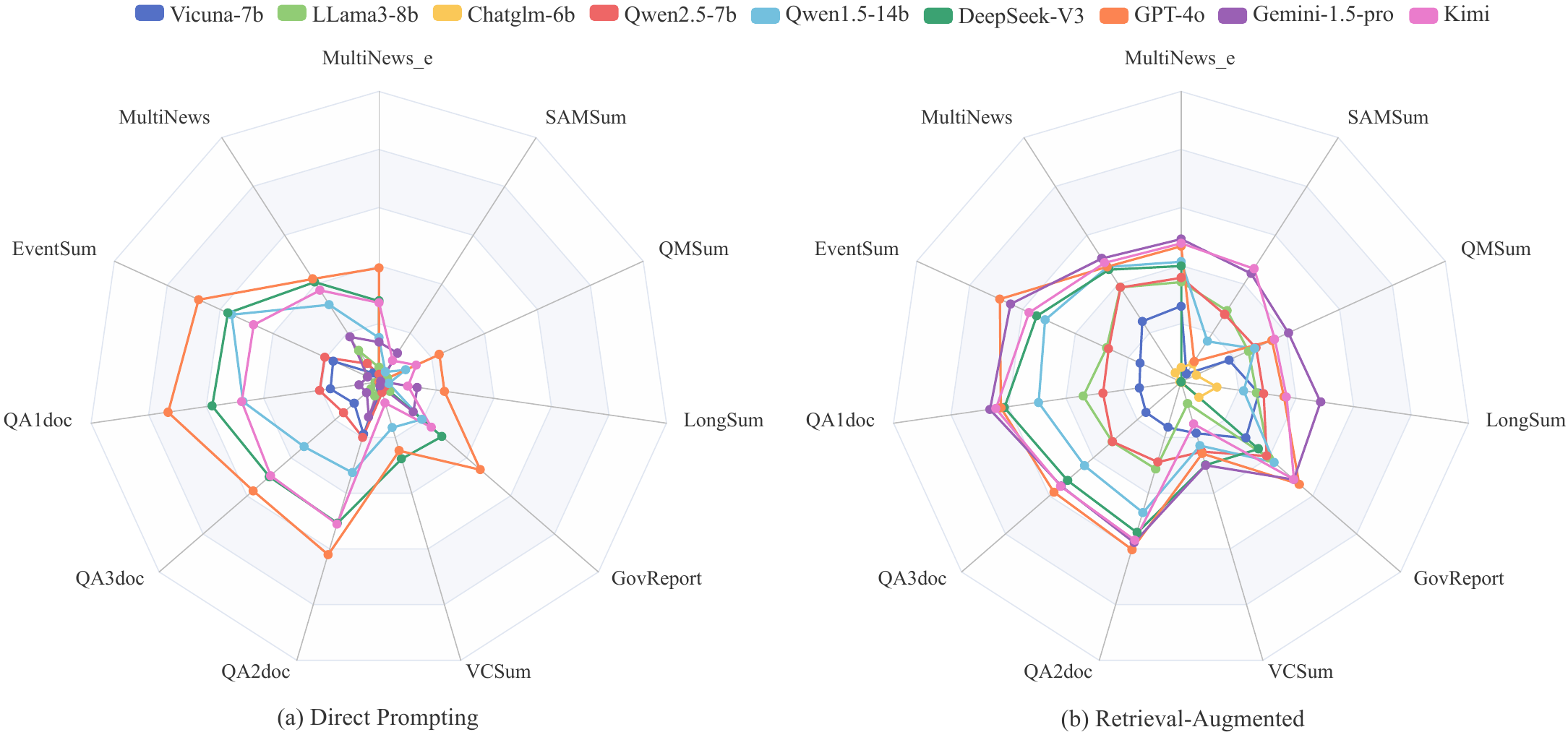}
    \caption{The performance of different models varies on different scenarios.}
    \label{fig:radar}
\end{figure*}

\begin{figure*}[t]
    \centering
	\setlength{\abovecaptionskip}{0.12cm}    
	\setlength{\belowcaptionskip}{-0.6cm}
	\includegraphics[width=0.92\linewidth]{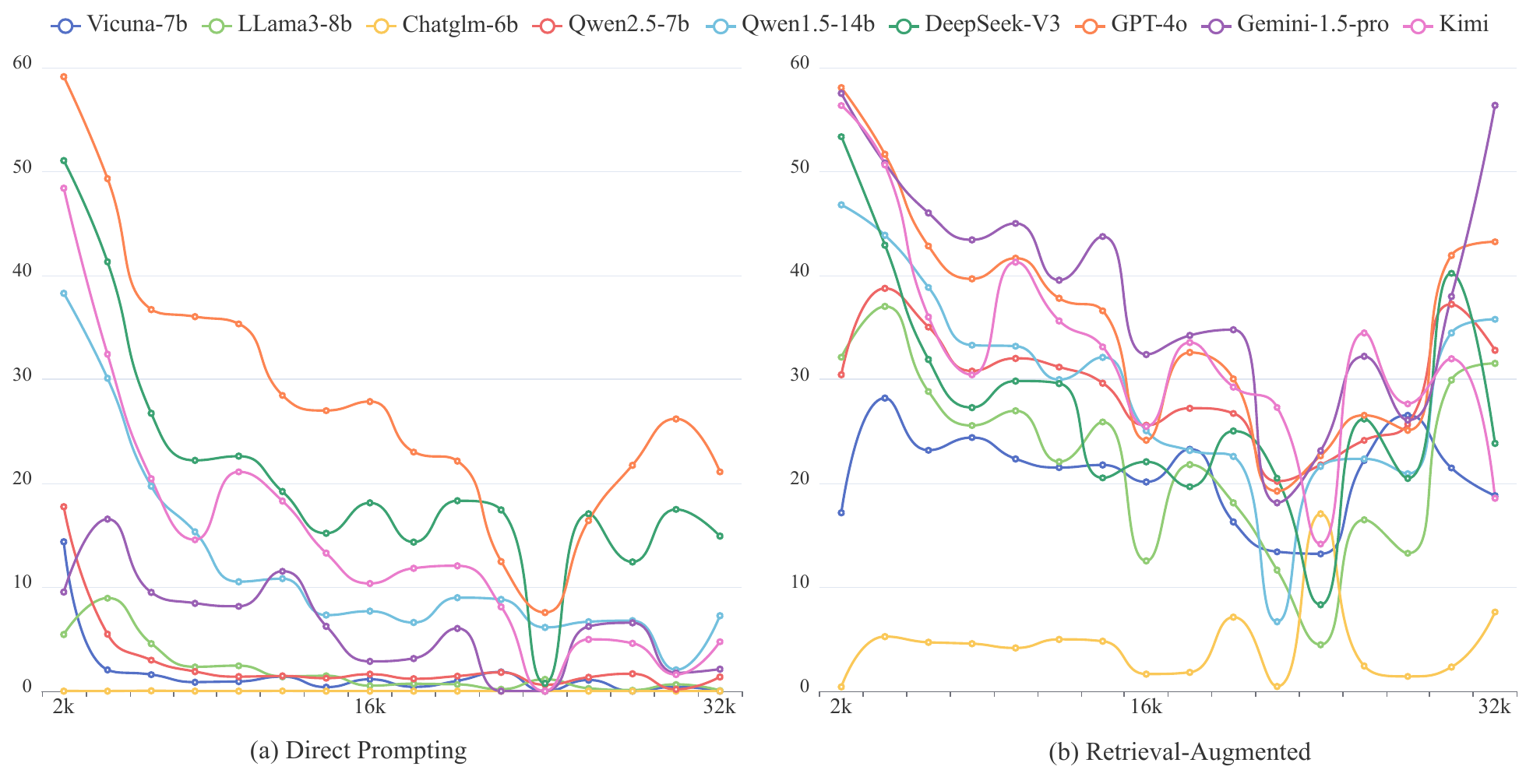}
    \caption{The performance of different models varies on different sourth lengths.}
    \label{fig:length}
\end{figure*}

\textbf{Relationship Classification}. Besides source sentence tracing, models must identify the relationship between traced and target sentences (e.g., quotation, compression, and inference). Relationship classification is more challenging than sentence tracing, requiring the model to understand deeper semantic and structural differences. Larger models (e.g., Qwen2.5-14B, Deepseek-V3, and GPT-4o) tend to perform more consistently, showing higher precision and recall in relationship classification than smaller open-source models. However, no model achieves highly reliable performance, suggesting that accurately capturing deep semantic relationships remains a challenging problem.

\begin{figure*}[t]
    \centering
	\setlength{\abovecaptionskip}{0.6cm}    
	\setlength{\belowcaptionskip}{-0.6cm}
    
    \begin{subfigure}[t]{0.49\textwidth}
        \centering
        \includegraphics[width=0.98\linewidth]{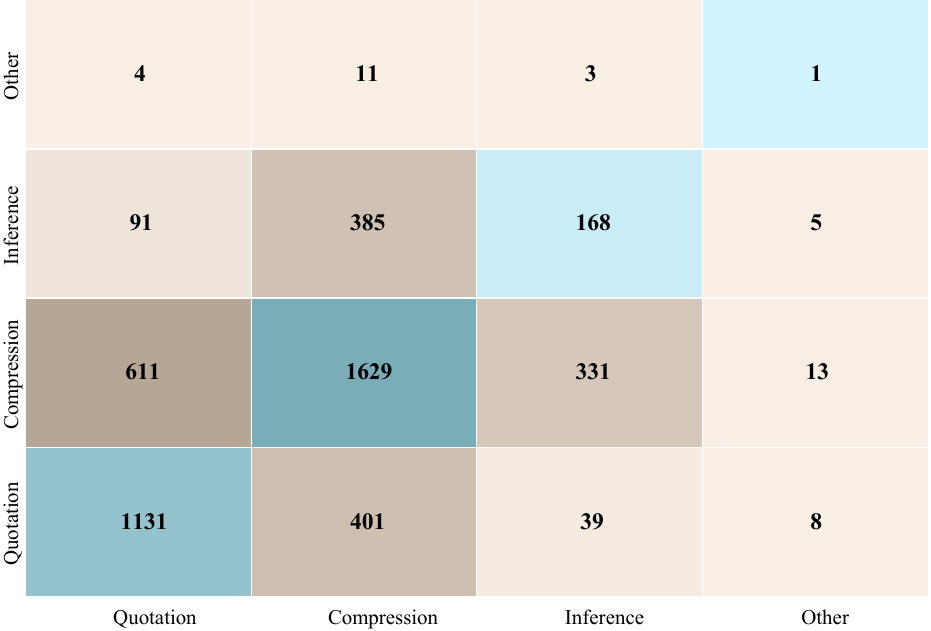}
        \caption{Confusion matrix for GPT-4o (DP).}
        \label{confusion-dp}
    \end{subfigure}
    \hfill 
    \begin{subfigure}[t]{0.49\textwidth}
        \centering
        \includegraphics[width=0.98\linewidth]{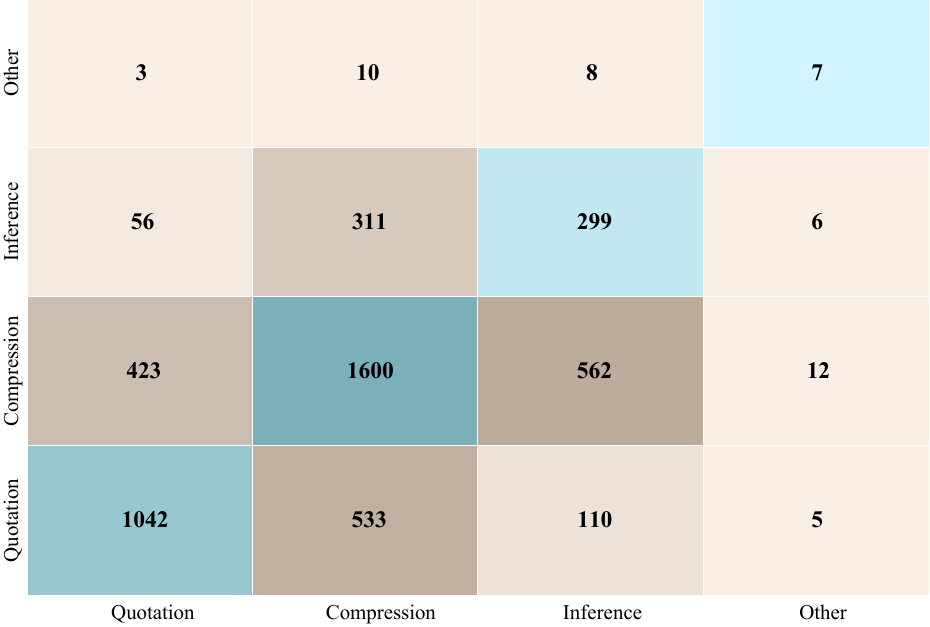}
        \caption{Confusion matrix for GPT-4o (RA).}
        \label{confusion-ra}
    \end{subfigure}
    
    \caption{Confusion matrix for GPT-4o. X-axis: predictions, y-axis: ground truth.}
    \label{confusion}
\end{figure*}

To conclude our analysis, we highlight the following key insights: 1) \textbf{Retrieval is essential.} Every model benefits significantly from retrieval, often turning poor performance in direct prompting into much stronger results when relevant context is provided. 2) \textbf{Larger models handle complex tasks better.} Larger models tend to perform better in relationship classification, indicating that richer representations are crucial for handling complex tasks. 3) \textbf{Precision and recall involve trade-offs.} Some models focus on capturing more potential sources, leading to higher recall but lower precision, while others do the opposite. The choice between high recall and high precision depends on the specific application. 4) \textbf{Closed-source models dominate, but open-source is catching up.} Models like Gemini-1.5-Pro and GPT-4o achieve the highest F1 scores, maintaining a clear advantage. However, retrieval-augmented open-source models, such as Qwen2.5-14B, are making significant progress and, in some cases, reaching comparable performance. 5) \textbf{Relationship classification remains a challenge.} No model achieves consistently strong performance in detecting complex relationships, showing that there is still room for improvement in fine-grained provenance tasks.

\subsection{Analysis}
\autoref{fig:radar} and \autoref{fig:length} show model performance across scenarios and source lengths respectively.

\textbf{Models Performance Across Scenarios.} Under the DP method, GPT-4o performs better than all other models across all scenarios. With the RA method, Kimi, GPT-4o, and Gemini-1.5-Pro each exhibit distinct advantages in different scenarios. For instance, GPT-4o leads in EventSum, QA3doc, and QA2doc; Gemini-1.5-Pro outperforms others in MultiNews, MultiNews\_e, QMSum, QA1doc, and LongSum; and Kimi shows outstanding performance in SAMSum. These results suggest that each model adapts differently depending on the scenario. They also show that RA leads to more significant improvements than DP, especially in multi-document, dialogue, and meeting scenarios.

\textbf{Models Performance Across Source Lengths.} Regarding source length, the RA is generally less affected by longer texts. In the DP, once the source length reaches 32k, only GPT-4o and DeepSeek-V3 maintain a passable but somewhat lower level of performance, while the others see a significant drop. Interestingly, Qwen2.5-14b usually falls behind Kimi and Gemini-1.5-Pro, but it surpasses both when the source length exceeds 20k.

\textbf{Error Analysis}. To understand model behavior, we analyze the confusion matrices for GPT-4o under both direct prompting and retrieval-augmented conditions, with results shown in \autoref{confusion-dp} and \autoref{confusion-ra} respectively. The matrices reveal a clear hierarchy of relationship difficulty: \textit{Quotation} is easiest, followed by \textit{Compression}, while \textit{Inference} proves most challenging. Three error patterns emerge: (1) Compression bias — models overpredict this category, with 533 \textit{Quotation} and 311 \textit{Inference} instances misclassified as \textit{Compression} in RA; (2) "Inference→Compression" confusion — 311 out of 672 true \textit{Inference} cases are misclassified as \textit{Compression}, indicating difficulty distinguishing between summarization and logical derivation; (3) "Other" underrepresentation — only 7 out of 28 instances correctly identified, highlighting challenges with rare relationship types.

Impact of Retrieval. While RA improves overall performance (\textit{Inference} correct predictions increase from 168 to 299), it also intensifies misclassification attempts. The Quotation→Compression errors increase from 401 in DP to 533 in RA, suggesting that additional context sometimes causes models to overinterpret simple quotations as more complex relationships. These patterns reveal fundamental challenges in relationship classification that extend beyond performance metrics.

\section{Conclusion}

We present TROVE, a fine-grained text provenance challenge to enhance transparency and accountability in text generation. TROVE traces each target sentence to its source, classifying their relationship as \emph{quotation}, \emph{compression}, \emph{inference}, or \emph{others}. TROVE offers a rigorous foundation for understanding where and how text is derived. Our dataset construction leverages three public datasets, LongBench, LooGLE, and CRUD-RAG, covering 11 scenarios, 2 languages, and 3 source length ranges.

Experiments with major LLMs show that retrieval augmentation significantly improves performance, especially for multi- and long-document settings. Larger models handle complex target-source relationships better, and while closed-source models lead in performance, open-source models reduce the gap with retrieval methods. However, relationship classification remains a key challenge.

\section*{Limitations}

We conclude the limitations of our study as follows: (1) Lack of Hallucination Cases. Our dataset construction relies on existing public datasets rather than texts generated directly by language models. As a result, hallucinations are absent in TROVE. In future work, we will enrich the dataset by incorporating model-generated content. (2) Scalability and Context Window Constraints. Although we include long-document and multi-document settings, current LLMs are constrained by finite context windows. In extremely lengthy documents, crucial source sentences might be ignored during retrieval. 


\section*{Acknowledgements}
The research work described in this paper has been partially supported by the Natural Science Foundation of China under Grant No. 62336008. We sincerely thank the anonymous reviewers for their insightful comments and constructive suggestions.

\bibliography{custom}

\clearpage
\appendix

\begin{table*}[t]
\centering
\setlength{\abovecaptionskip}{0.12cm}    
\setlength{\belowcaptionskip}{-0.2cm}
\begin{tabular}{@{}cc|ccccr@{}}
\toprule
\#Doc                   & Lang                & Tasks        & Number & Origin Dataset & Domain   & \#AvgLen \\ \midrule
\multirow{2}{*}{multi}  & \multirow{2}{*}{en} & MultiNews\_e & 133    & Long-Bench     & News     &  8,672.77            \\
                        &                     & MultiNews    & 47     & Long-Bench     & News     &  3,118.66            \\ \midrule
\multirow{4}{*}{single} & \multirow{4}{*}{en} & GovReport    & 63     & Long-Bench     & Report   & 6,836.30             \\
                        &                    & LongSum      & 15     & LooGLE         & ArXiv    & 21,797.40             \\
                        &                     & QMSum        & 47     & Long-Bench     & Meeting  & 9,222.79             \\
                        &                 & SAMSum       & 55     & Long-Bench     & Dialogue & 7,587.96             \\ \midrule
\multirow{2}{*}{multi}  & \multirow{2}{*}{zh} & QA2doc       & 90     & CRUD           & News     & 713.97             \\
                        &                     & QA3doc       & 90     & CRUD           & News     & 1,070.50            \\ \midrule
\multirow{3}{*}{single} & \multirow{3}{*}{zh} & EventSum     & 32     & CRUD           & News     & 758.97             \\
                        &                     & QA1doc       & 32     & CRUD           & News     & 676.41             \\
                        &                     & VCSum        & 116    & Long-Bench     & Meeting  & 11,993.00             \\ \bottomrule
\end{tabular}
\caption{Detailed statistics of our dataset. \#AvgLen denotes the average length of the source document(s), measured in Chinese characters for Chinese texts and words for English texts. \textit{Tasks} indicates the data's original task (scenario).}
\label{tab:datasets_appendix}
\end{table*}

\begin{table}[t]
\centering
\begin{tabular}{l|l}
\hline
Final-Lable & Pre-Label \\ \hline
\multirow{3}{*}{Quotion} & Copy \\
 & Paraphrase \\
 & Reordering \\ \hline
\multirow{3}{*}{Compression} & Fusion \\
 & Summary \\
 & Distillation \\ \hline
\multirow{3}{*}{Inference} & Inference \\
 & Expansion \\
 & Generalization \\ \hline
Other & Negation \\ \hline
\end{tabular}%
\caption{Mapping between pre-labels and final-labels.}
\label{tab:label_mapping}
\end{table}

\begin{table*}[t]
\centering
\setlength{\abovecaptionskip}{0.12cm}    
\setlength{\belowcaptionskip}{-0.6cm}
\resizebox{\textwidth}{!}{%
\begin{tabular}{@{}ccccrrrrrrrrrrrrr@{}}
\toprule
\multirow{2}{*}{LLM}                                                    & \multirow{2}{*}{\#Doc}  & \multirow{2}{*}{Lang.} & \multirow{2}{*}{Method} & \multicolumn{6}{c}{Macro}                                                                                                                                          & \multicolumn{6}{c}{Micro}                                                                                                                                          & \multicolumn{1}{c}{\multirow{2}{*}{F1}} \\ \cmidrule(lr){5-16}
                                                                        &                         &                        &                         & \multicolumn{1}{c}{T\_P} & \multicolumn{1}{c}{T\_R} & \multicolumn{1}{l}{T\_F1} & \multicolumn{1}{c}{R\_P} & \multicolumn{1}{c}{R\_R} & \multicolumn{1}{c|}{R\_F1} & \multicolumn{1}{c}{T\_P} & \multicolumn{1}{c}{T\_R} & \multicolumn{1}{l}{T\_F1} & \multicolumn{1}{c}{R\_P} & \multicolumn{1}{c}{R\_R} & \multicolumn{1}{c|}{R\_F1} & \multicolumn{1}{c}{}                    \\ \midrule
\multirow{8}{*}{Vicuna}                                                 & \multirow{4}{*}{single} & \multirow{2}{*}{en}    & \multicolumn{1}{c|}{DP} & 0.45                     & 2.41                     & 0.76                      & 0.21                     & 1.52                     & \multicolumn{1}{r|}{0.37}  & 0.42                     & 2.47                     & 0.72                      & 0.19                     & 1.43                     & \multicolumn{1}{r|}{0.34}  & 0.55                                    \\
                                                                        &                         &                        & \multicolumn{1}{c|}{RA} & 29.82                    & 46.96                    & 36.48                     & 10.50                    & 19.69                    & \multicolumn{1}{r|}{13.69} & 32.01                    & 45.45                    & 37.56                     & 11.05                    & 16.64                    & \multicolumn{1}{r|}{13.28} & 25.25                                   \\
                                                                        &                         & \multirow{2}{*}{zh}    & \multicolumn{1}{c|}{DP} & 8.97                     & 30.53                    & 13.86                     & 3.19                     & 10.94                    & \multicolumn{1}{r|}{4.94}  & 8.52                     & 30.51                    & 13.33                     & 3.04                     & 11.08                    & \multicolumn{1}{r|}{4.77}  & 9.22                                    \\
                                                                        &                         &                        & \multicolumn{1}{c|}{RA} & 29.75                    & 32.47                    & 31.05                     & 10.12                    & 12.75                    & \multicolumn{1}{r|}{11.28} & 30.34                    & 32.81                    & 31.53                     & 10.26                    & 12.63                    & \multicolumn{1}{r|}{11.33} & 21.30                                   \\ \cmidrule(l){2-17} 
                                                                        & \multirow{4}{*}{multi}  & \multirow{2}{*}{en}    & \multicolumn{1}{c|}{DP} & 2.45                     & 10.85                    & 4.00                      & 0.82                     & 3.64                     & \multicolumn{1}{r|}{1.33}  & 2.21                     & 10.70                    & 3.66                      & 0.68                     & 3.55                     & \multicolumn{1}{r|}{1.15}  & 2.54                                    \\
                                                                        &                         &                        & \multicolumn{1}{c|}{RA} & 30.89                    & 48.43                    & 37.72                     & 11.53                    & 21.08                    & \multicolumn{1}{r|}{14.91} & 34.62                    & 46.45                    & 39.68                     & 12.84                    & 18.39                    & \multicolumn{1}{r|}{15.12} & 26.86                                   \\
                                                                        &                         & \multirow{2}{*}{zh}    & \multicolumn{1}{c|}{DP} & 15.33                    & 49.10                    & 23.37                     & 5.25                     & 18.01                    & \multicolumn{1}{r|}{8.13}  & 15.96                    & 48.70                    & 24.04                     & 5.60                     & 17.66                    & \multicolumn{1}{r|}{8.50}  & 16.01                                   \\
                                                                        &                         &                        & \multicolumn{1}{c|}{RA} & 18.09                    & 39.17                    & 24.75                     & 7.03                     & 15.34                    & \multicolumn{1}{r|}{9.64}  & 19.43                    & 38.40                    & 25.81                     & 7.42                     & 15.39                    & \multicolumn{1}{r|}{10.02} & 17.55                                   \\ \midrule
\multirow{8}{*}{LLama3}                                                 & \multirow{4}{*}{single} & \multirow{2}{*}{en}    & \multicolumn{1}{c|}{DP} & 3.53                     & 30.86                    & 6.33                      & 1.37                     & 12.05                    & \multicolumn{1}{r|}{2.45}  & 2.55                     & 29.67                    & 4.69                      & 0.95                     & 10.75                    & \multicolumn{1}{r|}{1.75}  & 3.81                                    \\
                                                                        &                         &                        & \multicolumn{1}{c|}{RA} & 55.74                    & 43.15                    & 48.64                     & 24.80                    & 18.66                    & \multicolumn{1}{r|}{21.30} & 58.79                    & 39.01                    & 46.90                     & 25.79                    & 17.86                    & \multicolumn{1}{r|}{21.10} & 34.49                                   \\
                                                                        &                         & \multirow{2}{*}{zh}    & \multicolumn{1}{c|}{DP} & 1.51                     & 1.70                     & 1.60                      & 0.52                     & 0.65                     & \multicolumn{1}{r|}{0.58}  & 2.10                     & 1.46                     & 1.72                      & 0.59                     & 0.45                     & \multicolumn{1}{r|}{0.51}  & 1.10                                    \\
                                                                        &                         &                        & \multicolumn{1}{c|}{RA} & 27.66                    & 20.11                    & 23.29                     & 14.95                    & 11.26                    & \multicolumn{1}{r|}{12.84} & 32.05                    & 19.55                    & 24.28                     & 17.24                    & 11.30                    & \multicolumn{1}{r|}{13.65} & 18.52                                   \\ \cmidrule(l){2-17} 
                                                                        & \multirow{4}{*}{multi}  & \multirow{2}{*}{en}    & \multicolumn{1}{c|}{DP} & 8.24                     & 26.55                    & 12.57                     & 4.57                     & 13.90                    & \multicolumn{1}{r|}{6.88}  & 7.12                     & 23.76                    & 10.96                     & 3.75                     & 11.58                    & \multicolumn{1}{r|}{5.67}  & 9.02                                    \\
                                                                        &                         &                        & \multicolumn{1}{c|}{RA} & 49.07                    & 47.30                    & 48.17                     & 26.08                    & 25.94                    & \multicolumn{1}{r|}{26.01} & 55.06                    & 42.50                    & 47.97                     & 29.72                    & 23.10                    & \multicolumn{1}{r|}{26.00} & 37.04                                   \\
                                                                        &                         & \multirow{2}{*}{zh}    & \multicolumn{1}{c|}{DP} & 7.37                     & 7.42                     & 7.39                      & 1.67                     & 2.25                     & \multicolumn{1}{r|}{1.92}  & 10.04                    & 7.13                     & 8.34                      & 2.05                     & 2.03                     & \multicolumn{1}{r|}{2.04}  & 4.92                                    \\
                                                                        &                         &                        & \multicolumn{1}{c|}{RA} & 42.50                    & 42.18                    & 42.34                     & 24.14                    & 22.19                    & \multicolumn{1}{r|}{23.13} & 53.36                    & 39.13                    & 45.15                     & 28.94                    & 21.35                    & \multicolumn{1}{r|}{24.57} & 33.80                                   \\ \midrule
\multirow{8}{*}{Chatglm}                                                & \multirow{4}{*}{single} & \multirow{2}{*}{en}    & \multicolumn{1}{c|}{DP} & 0.00                     & 0.00                     & 0.00                      & 0.00                     & 0.00                     & \multicolumn{1}{r|}{0.00}  & 0.00                     & 0.00                     & 0.00                      & 0.00                     & 0.00                     & \multicolumn{1}{r|}{0.00}  & 0.00                                    \\
                                                                        &                         &                        & \multicolumn{1}{c|}{RA} & 8.95                     & 9.73                     & 9.32                      & 3.80                     & 4.58                     & \multicolumn{1}{r|}{4.16}  & 25.06                    & 9.92                     & 14.21                     & 11.35                    & 4.70                     & \multicolumn{1}{r|}{6.65}  & 8.58                                    \\
                                                                        &                         & \multirow{2}{*}{zh}    & \multicolumn{1}{c|}{DP} & 0.00                     & 0.00                     & 0.00                      & 0.00                     & 0.00                     & \multicolumn{1}{r|}{0.00}  & 0.00                     & 0.00                     & 0.00                      & 0.00                     & 0.00                     & \multicolumn{1}{r|}{0.00}  & 0.00                                    \\
                                                                        &                         &                        & \multicolumn{1}{c|}{RA} & 0.42                     & 0.20                     & 0.27                      & 0.28                     & 0.06                     & \multicolumn{1}{r|}{0.10}  & 0.83                     & 0.17                     & 0.29                      & 0.28                     & 0.06                     & \multicolumn{1}{r|}{0.10}  & 0.19                                    \\ \cmidrule(l){2-17} 
                                                                        & \multirow{4}{*}{multi}  & \multirow{2}{*}{en}    & \multicolumn{1}{c|}{DP} & 0.06                     & 0.17                     & 0.09                      & 0.04                     & 0.01                     & \multicolumn{1}{r|}{0.02}  & 0.16                     & 0.08                     & 0.11                      & 0.02                     & 0.02                     & \multicolumn{1}{r|}{0.02}  & 0.06                                    \\
                                                                        &                         &                        & \multicolumn{1}{c|}{RA} & 5.32                     & 6.23                     & 5.74                      & 1.91                     & 2.39                     & \multicolumn{1}{r|}{2.13}  & 21.90                    & 5.79                     & 9.15                      & 7.75                     & 2.15                     & \multicolumn{1}{r|}{3.37}  & 5.10                                    \\
                                                                        &                         & \multirow{2}{*}{zh}    & \multicolumn{1}{c|}{DP} & 0.00                     & 0.00                     & 0.00                      & 0.00                     & 0.00                     & \multicolumn{1}{r|}{0.00}  & 0.00                     & 0.00                     & 0.00                      & 0.00                     & 0.00                     & \multicolumn{1}{r|}{0.00}  & 0.00                                    \\
                                                                        &                         &                        & \multicolumn{1}{c|}{RA} & 0.02                     & 0.07                     & 0.04                      & 0.00                     & 0.00                     & \multicolumn{1}{r|}{0.00}  & 0.10                     & 0.05                     & 0.07                      & 0.00                     & 0.00                     & \multicolumn{1}{r|}{0.00}  & 0.03                                    \\ \midrule
\multirow{8}{*}{\begin{tabular}[c]{@{}c@{}}Qwen1.5\\ -7b\end{tabular}}  & \multirow{4}{*}{single} & \multirow{2}{*}{en}    & \multicolumn{1}{c|}{DP} & 1.30                     & 30.84                    & 2.50                      & 0.39                     & 12.05                    & \multicolumn{1}{r|}{0.76}  & 1.29                     & 30.19                    & 2.47                      & 0.40                     & 9.52                     & \multicolumn{1}{r|}{0.76}  & 1.62                                    \\
                                                                        &                         &                        & \multicolumn{1}{c|}{RA} & 33.91                    & 57.06                    & 42.54                     & 11.07                    & 22.61                    & \multicolumn{1}{r|}{14.87} & 31.66                    & 55.25                    & 40.26                     & 9.96                     & 18.58                    & \multicolumn{1}{r|}{12.97} & 27.66                                   \\
                                                                        &                         & \multirow{2}{*}{zh}    & \multicolumn{1}{c|}{DP} & 7.22                     & 37.80                    & 12.13                     & 1.91                     & 10.51                    & \multicolumn{1}{r|}{3.23}  & 7.07                     & 37.76                    & 11.91                     & 1.87                     & 10.20                    & \multicolumn{1}{r|}{3.16}  & 7.61                                    \\
                                                                        &                         &                        & \multicolumn{1}{c|}{RA} & 43.16                    & 42.11                    & 42.63                     & 10.25                    & 11.65                    & \multicolumn{1}{r|}{10.90} & 42.95                    & 42.11                    & 42.53                     & 10.01                    & 11.02                    & \multicolumn{1}{r|}{10.49} & 26.64                                   \\ \cmidrule(l){2-17} 
                                                                        & \multirow{4}{*}{multi}  & \multirow{2}{*}{en}    & \multicolumn{1}{c|}{DP} & 2.04                     & 29.08                    & 3.81                      & 0.48                     & 8.94                     & \multicolumn{1}{r|}{0.91}  & 1.88                     & 28.85                    & 3.53                      & 0.48                     & 7.17                     & \multicolumn{1}{r|}{0.91}  & 2.29                                    \\
                                                                        &                         &                        & \multicolumn{1}{c|}{RA} & 36.45                    & 54.01                    & 43.52                     & 14.62                    & 24.34                    & \multicolumn{1}{r|}{18.27} & 34.91                    & 50.12                    & 41.15                     & 13.16                    & 19.59                    & \multicolumn{1}{r|}{15.75} & 29.67                                   \\
                                                                        &                         & \multirow{2}{*}{zh}    & \multicolumn{1}{c|}{DP} & 16.40                    & 66.28                    & 26.30                     & 2.67                     & 15.33                    & \multicolumn{1}{r|}{4.54}  & 15.75                    & 66.48                    & 25.46                     & 2.34                     & 13.28                    & \multicolumn{1}{r|}{3.98}  & 15.07                                   \\
                                                                        &                         &                        & \multicolumn{1}{c|}{RA} & 30.44                    & 59.81                    & 40.35                     & 8.13                     & 17.71                    & \multicolumn{1}{r|}{11.15} & 29.79                    & 61.26                    & 40.08                     & 7.89                     & 16.73                    & \multicolumn{1}{r|}{10.72} & 25.58                                   \\ \midrule
\multirow{8}{*}{\begin{tabular}[c]{@{}c@{}}Qwen2.5\\ -7b\end{tabular}}  & \multirow{4}{*}{single} & \multirow{2}{*}{en}    & \multicolumn{1}{c|}{DP} & 1.41                     & 25.92                    & 2.68                      & 0.67                     & 9.42                     & \multicolumn{1}{r|}{1.25}  & 1.03                     & 25.63                    & 1.97                      & 0.29                     & 6.95                     & \multicolumn{1}{r|}{0.55}  & 1.61                                    \\
                                                                        &                         &                        & \multicolumn{1}{c|}{RA} & 44.35                    & 67.54                    & 53.54                     & 15.43                    & 28.65                    & \multicolumn{1}{r|}{20.06} & 41.56                    & 64.62                    & 50.58                     & 12.21                    & 22.88                    & \multicolumn{1}{r|}{15.92} & 35.02                                   \\
                                                                        &                         & \multirow{2}{*}{zh}    & \multicolumn{1}{c|}{RA} & 10.38                    & 46.09                    & 16.95                     & 3.54                     & 14.21                    & \multicolumn{1}{r|}{5.67}  & 9.30                     & 46.08                    & 15.48                     & 2.68                     & 13.32                    & \multicolumn{1}{r|}{4.46}  & 10.64                                   \\
                                                                        &                         &                        & \multicolumn{1}{c|}{RA} & 43.67                    & 59.50                    & 50.37                     & 12.06                    & 20.57                    & \multicolumn{1}{r|}{15.20} & 42.69                    & 59.01                    & 49.54                     & 11.32                    & 19.68                    & \multicolumn{1}{r|}{14.38} & 32.37                                   \\ \cmidrule(l){2-17} 
                                                                        & \multirow{4}{*}{multi}  & \multirow{2}{*}{en}    & \multicolumn{1}{c|}{DP} & 4.36                     & 40.63                    & 7.87                      & 2.27                     & 14.96                    & \multicolumn{1}{r|}{3.94}  & 2.59                     & 40.36                    & 4.87                      & 0.72                     & 12.24                    & \multicolumn{1}{r|}{1.35}  & 4.51                                    \\
                                                                        &                         &                        & \multicolumn{1}{c|}{RA} & 45.50                    & 70.35                    & 55.26                     & 20.70                    & 33.92                    & \multicolumn{1}{r|}{25.71} & 41.16                    & 66.15                    & 50.74                     & 16.45                    & 27.40                    & \multicolumn{1}{r|}{20.55} & 38.07                                   \\
                                                                        &                         & \multirow{2}{*}{zh}    & \multicolumn{1}{c|}{DP} & 19.36                    & 85.59                    & 31.58                     & 5.10                     & 23.38                    & \multicolumn{1}{r|}{8.37}  & 17.75                    & 85.93                    & 29.43                     & 4.03                     & 21.75                    & \multicolumn{1}{r|}{6.80}  & 19.04                                   \\
                                                                        &                         &                        & \multicolumn{1}{c|}{RA} & 34.32                    & 90.28                    & 49.73                     & 10.73                    & 31.09                    & \multicolumn{1}{r|}{15.96} & 32.60                    & 89.09                    & 47.74                     & 9.29                     & 28.26                    & \multicolumn{1}{r|}{13.98} & 31.85                                   \\ \midrule
\multirow{8}{*}{\begin{tabular}[c]{@{}c@{}}Qwen1.5\\ -14b\end{tabular}} & \multirow{4}{*}{single} & \multirow{2}{*}{en}    & \multicolumn{1}{c|}{DP} & 0.77                     & 31.02                    & 1.50                      & 0.17                     & 7.48                     & \multicolumn{1}{r|}{0.33}  & 0.72                     & 27.75                    & 1.41                      & 0.13                     & 5.76                     & \multicolumn{1}{r|}{0.26}  & 0.88                                    \\
                                                                        &                         &                        & \multicolumn{1}{c|}{RA} & 46.18                    & 44.82                    & 45.49                     & 9.95                     & 14.05                    & \multicolumn{1}{r|}{11.65} & 50.84                    & 43.02                    & 46.61                     & 12.25                    & 12.13                    & \multicolumn{1}{r|}{12.19} & 28.99                                   \\
                                                                        &                         & \multirow{2}{*}{zh}    & \multicolumn{1}{c|}{DP} & 15.79                    & 32.88                    & 21.34                     & 4.84                     & 10.01                    & \multicolumn{1}{r|}{6.52}  & 15.73                    & 32.76                    & 21.26                     & 4.46                     & 9.54                     & \multicolumn{1}{r|}{6.08}  & 13.80                                   \\
                                                                        &                         &                        & \multicolumn{1}{c|}{RA} & 45.66                    & 41.57                    & 43.52                     & 11.60                    & 14.17                    & \multicolumn{1}{r|}{12.75} & 47.04                    & 41.15                    & 43.90                     & 12.02                    & 13.12                    & \multicolumn{1}{r|}{12.55} & 28.18                                   \\ \cmidrule(l){2-17} 
                                                                        & \multirow{4}{*}{multi}  & \multirow{2}{*}{en}    & \multicolumn{1}{c|}{DP} & 6.37                     & 48.85                    & 11.27                     & 2.85                     & 22.98                    & \multicolumn{1}{r|}{5.07}  & 4.38                     & 46.69                    & 8.01                      & 1.58                     & 18.51                    & \multicolumn{1}{r|}{2.92}  & 6.82                                    \\
                                                                        &                         &                        & \multicolumn{1}{c|}{RA} & 50.35                    & 64.90                    & 56.71                     & 23.14                    & 32.69                    & \multicolumn{1}{r|}{27.10} & 49.75                    & 60.59                    & 54.64                     & 20.70                    & 26.41                    & \multicolumn{1}{r|}{23.21} & 40.41                                   \\
                                                                        &                         & \multirow{2}{*}{zh}    & \multicolumn{1}{c|}{DP} & 27.05                    & 43.14                    & 33.25                     & 6.99                     & 12.95                    & \multicolumn{1}{r|}{9.08}  & 32.60                    & 42.85                    & 37.03                     & 8.44                     & 11.46                    & \multicolumn{1}{r|}{9.72}  & 22.27                                   \\
                                                                        &                         &                        & \multicolumn{1}{c|}{RA} & 38.09                    & 63.45                    & 47.61                     & 11.57                    & 22.93                    & \multicolumn{1}{r|}{15.38} & 44.33                    & 61.84                    & 51.64                     & 13.02                    & 19.92                    & \multicolumn{1}{r|}{15.75} & 32.59                                   \\ \midrule
\multirow{8}{*}{\begin{tabular}[c]{@{}c@{}}Qwen2.5\\ -14b\end{tabular}} & \multirow{4}{*}{single} & \multirow{2}{*}{en}    & \multicolumn{1}{c|}{DP} & 15.92                    & 49.57                    & 24.10                     & 9.88                     & 25.78                    & \multicolumn{1}{r|}{14.29} & 7.90                     & 45.88                    & 13.48                     & 4.40                     & 22.10                    & \multicolumn{1}{r|}{7.34}  & 14.80                                   \\
                                                                        &                         &                        & \multicolumn{1}{c|}{RA} & 44.43                    & 42.18                    & 43.28                     & 24.75                    & 23.71                    & \multicolumn{1}{r|}{24.22} & 60.35                    & 37.72                    & 46.43                     & 31.53                    & 19.50                    & \multicolumn{1}{r|}{24.10} & 34.50                                   \\
                                                                        &                         & \multirow{2}{*}{zh}    & \multicolumn{1}{c|}{DP} & 36.61                    & 50.26                    & 42.36                     & 17.77                    & 25.25                    & \multicolumn{1}{r|}{20.86} & 36.70                    & 49.58                    & 42.18                     & 17.12                    & 24.10                    & \multicolumn{1}{r|}{20.02} & 31.35                                   \\
                                                                        &                         &                        & \multicolumn{1}{c|}{RA} & 57.24                    & 39.46                    & 46.71                     & 29.73                    & 21.04                    & \multicolumn{1}{r|}{24.64} & 62.04                    & 39.30                    & 48.12                     & 31.79                    & 20.85                    & \multicolumn{1}{r|}{25.18} & 36.16                                   \\ \cmidrule(l){2-17} 
                                                                        & \multirow{4}{*}{multi}  & \multirow{2}{*}{en}    & \multicolumn{1}{c|}{DP} & 21.86                    & 62.59                    & 32.40                     & 12.43                    & 36.63                    & \multicolumn{1}{r|}{18.56} & 16.92                    & 57.50                    & 26.14                     & 8.92                     & 31.71                    & \multicolumn{1}{r|}{13.92} & 22.76                                   \\
                                                                        &                         &                        & \multicolumn{1}{c|}{RA} & 53.62                    & 57.14                    & 55.33                     & 31.55                    & 34.13                    & \multicolumn{1}{r|}{32.79} & 65.04                    & 51.61                    & 57.55                     & 36.25                    & 29.43                    & \multicolumn{1}{r|}{32.49} & 44.54                                   \\
                                                                        &                         & \multirow{2}{*}{zh}    & \multicolumn{1}{c|}{DP} & 44.69                    & 46.59                    & 45.62                     & 20.37                    & 21.81                    & \multicolumn{1}{r|}{21.06} & 56.50                    & 45.24                    & 50.25                     & 27.04                    & 21.07                    & \multicolumn{1}{r|}{23.69} & 35.16                                   \\
                                                                        &                         &                        & \multicolumn{1}{c|}{RA} & 63.09                    & 62.18                    & 62.63                     & 30.86                    & 30.84                    & \multicolumn{1}{r|}{30.85} & 71.27                    & 59.46                    & 64.83                     & 34.40                    & 29.42                    & \multicolumn{1}{r|}{31.72} & 47.51                                   \\ \midrule
\multirow{8}{*}{DeepSeek-V3}                                            & \multirow{4}{*}{single} & \multirow{2}{*}{en}    & \multicolumn{1}{c|}{DP} & 17.34                    & 23.46                    & 19.94                     & 11.26                    & 15.33                    & \multicolumn{1}{r|}{12.98} & 11.53                    & 21.44                    & 15.00                     & 6.83                     & 12.31                    & \multicolumn{1}{r|}{8.78}  & 14.17                                   \\
                                                                        &                         &                        & \multicolumn{1}{c|}{RA} & 18.22                    & 25.15                    & 21.13                     & 10.56                    & 16.07                    & \multicolumn{1}{r|}{12.75} & 20.07                    & 23.07                    & 21.47                     & 11.27                    & 13.24                    & \multicolumn{1}{r|}{12.18} & 16.88                                   \\
                                                                        &                         & \multirow{2}{*}{zh}    & \multicolumn{1}{c|}{DP} & 51.57                    & 62.88                    & 56.67                     & 23.24                    & 29.34                    & \multicolumn{1}{r|}{25.94} & 50.08                    & 62.59                    & 55.64                     & 21.56                    & 28.57                    & \multicolumn{1}{r|}{24.58} & 40.71                                   \\
                                                                        &                         &                        & \multicolumn{1}{c|}{RA} & 63.86                    & 56.21                    & 59.79                     & 31.08                    & 28.85                    & \multicolumn{1}{r|}{29.92} & 63.37                    & 55.65                    & 59.26                     & 30.15                    & 28.16                    & \multicolumn{1}{r|}{29.12} & 44.52                                   \\ \cmidrule(l){2-17} 
                                                                        & \multirow{4}{*}{multi}  & \multirow{2}{*}{en}    & \multicolumn{1}{c|}{DP} & 39.48                    & 60.90                    & 47.91                     & 23.39                    & 35.69                    & \multicolumn{1}{r|}{28.26} & 26.40                    & 56.82                    & 36.05                     & 14.01                    & 30.79                    & \multicolumn{1}{r|}{19.26} & 32.87                                   \\
                                                                        &                         &                        & \multicolumn{1}{c|}{RA} & 48.45                    & 59.18                    & 53.28                     & 29.25                    & 36.83                    & \multicolumn{1}{r|}{32.61} & 55.20                    & 55.54                    & 55.37                     & 31.88                    & 32.74                    & \multicolumn{1}{r|}{32.30} & 43.39                                   \\
                                                                        &                         & \multirow{2}{*}{zh}    & \multicolumn{1}{c|}{DP} & 70.78                    & 79.02                    & 74.67                     & 29.63                    & 33.24                    & \multicolumn{1}{r|}{31.33} & 70.15                    & 76.37                    & 73.13                     & 27.45                    & 30.77                    & \multicolumn{1}{r|}{29.02} & 52.04                                   \\
                                                                        &                         &                        & \multicolumn{1}{c|}{RA} & 66.13                    & 82.86                    & 73.56                     & 33.52                    & 43.04                    & \multicolumn{1}{r|}{37.69} & 64.37                    & 80.78                    & 71.65                     & 31.39                    & 40.01                    & \multicolumn{1}{r|}{35.18} & 54.52                                   \\ \bottomrule
\end{tabular}
}
\caption{Experiment results of open-source LLMs under single- and multi-document settings in English and Chinese. DP and RA denote direct prompting tracing and retrieval-augmented tracing.}
\label{tab:experiment-appendix1}
\end{table*}

\begin{table*}[t]
\centering
\setlength{\abovecaptionskip}{0.12cm}    
\setlength{\belowcaptionskip}{-0.6cm}
\resizebox{\textwidth}{!}{%
\begin{tabular}{cccc|rrrrrr|rrrrrr|r}
\toprule
\multirow{2}{*}{LLM}                                                       & \multirow{2}{*}{\#Doc}  & \multirow{2}{*}{Lang.} & \multirow{2}{*}{Method} & \multicolumn{6}{c}{Macro}                                                                                                                                          & \multicolumn{6}{c}{Micro}                                                                                                                                          & \multicolumn{1}{c}{\multirow{2}{*}{F1}} \\ \cmidrule(lr){5-16}
                                                                           &                         &                        &                         & \multicolumn{1}{c}{T\_P} & \multicolumn{1}{c}{T\_R} & \multicolumn{1}{l}{T\_F1} & \multicolumn{1}{c}{R\_P} & \multicolumn{1}{c}{R\_R} & \multicolumn{1}{c|}{R\_F1} & \multicolumn{1}{c}{T\_P} & \multicolumn{1}{c}{T\_R} & \multicolumn{1}{l}{T\_F1} & \multicolumn{1}{c}{R\_P} & \multicolumn{1}{c}{R\_R} & \multicolumn{1}{c|}{R\_F1} & \multicolumn{1}{c}{}                    \\ \midrule
\multirow{8}{*}{GPT-4o}                                                    & \multirow{4}{*}{single} & \multirow{2}{*}{en}    & \multicolumn{1}{c|}{DP} & 46.54                    & 41.11                    & 43.66                     & 28.69                    & 25.08                    & \multicolumn{1}{r|}{26.76} & 46.72                    & 37.16                    & 41.40                     & 27.16                    & 21.68                    & \multicolumn{1}{r|}{24.11} & 33.98                                   \\
                                                                           &                         &                        & \multicolumn{1}{c|}{RA} & 65.96                    & 50.93                    & 57.48                     & 35.99                    & 27.69                    & \multicolumn{1}{r|}{31.29} & 70.60                    & 46.50                    & 56.07                     & 36.01                    & 24.30                    & \multicolumn{1}{r|}{29.02} & 43.47                                   \\
                                                                           &                         & \multirow{2}{*}{zh}    & \multicolumn{1}{c|}{DP} & 59.17                    & 48.70                    & 53.43                     & 38.02                    & 31.44                    & \multicolumn{1}{r|}{34.42} & 58.57                    & 48.25                    & 52.91                     & 37.27                    & 30.71                    & \multicolumn{1}{r|}{33.68} & 43.61                                   \\
                                                                           &                         &                        & \multicolumn{1}{c|}{RA} & 75.47                    & 43.72                    & 55.36                     & 41.82                    & 24.84                    & \multicolumn{1}{r|}{31.17} & 75.76                    & 42.81                    & 54.71                     & 41.70                    & 24.32                    & \multicolumn{1}{r|}{30.73} & 42.99                                   \\ \cmidrule(l){2-17} 
                                                                           & \multirow{4}{*}{multi}  & \multirow{2}{*}{en}    & \multicolumn{1}{c|}{DP} & 50.62                    & 55.20                    & 52.81                     & 33.67                    & 35.36                    & \multicolumn{1}{r|}{34.49} & 45.03                    & 50.64                    & 47.67                     & 28.38                    & 31.93                    & \multicolumn{1}{r|}{30.05} & 41.26                                   \\
                                                                           &                         &                        & \multicolumn{1}{c|}{RA} & 64.98                    & 58.04                    & 61.31                     & 39.79                    & 36.00                    & \multicolumn{1}{r|}{37.80} & 67.07                    & 52.12                    & 58.66                     & 40.26                    & 31.85                    & \multicolumn{1}{r|}{35.57} & 48.34                                   \\
                                                                           &                         & \multirow{2}{*}{zh}    & \multicolumn{1}{c|}{DP} & 80.90                    & 76.83                    & 78.81                     & 45.83                    & 44.06                    & \multicolumn{1}{r|}{44.93} & 78.95                    & 73.67                    & 76.22                     & 44.75                    & 42.11                    & \multicolumn{1}{r|}{43.39} & 60.84                                   \\
                                                                           &                         &                        & \multicolumn{1}{c|}{RA} & 86.15                    & 69.12                    & 76.70                     & 53.14                    & 42.96                    & \multicolumn{1}{r|}{47.51} & 85.81                    & 64.93                    & 73.92                     & 53.76                    & 40.51                    & \multicolumn{1}{r|}{46.20} & 61.09                                   \\ \midrule
\multirow{8}{*}{\begin{tabular}[c]{@{}c@{}}Gemini-1.5\\ -pro\end{tabular}} & \multirow{4}{*}{single} & \multirow{2}{*}{en}    & \multicolumn{1}{c|}{DP} & 16.01                    & 13.85                    & 14.86                     & 8.83                     & 7.40                     & \multicolumn{1}{r|}{8.05}  & 12.85                    & 11.57                    & 12.17                     & 6.53                     & 5.96                     & \multicolumn{1}{r|}{6.23}  & 10.33                                   \\
                                                                           &                         &                        & \multicolumn{1}{c|}{RA} & 69.39                    & 55.51                    & 61.68                     & 42.20                    & 31.83                    & \multicolumn{1}{r|}{36.28} & 70.53                    & 50.68                    & 58.98                     & 44.17                    & 31.30                    & \multicolumn{1}{r|}{36.64} & 48.39                                   \\
                                                                           &                         & \multirow{2}{*}{zh}    & \multicolumn{1}{c|}{DP} & 4.34                     & 3.81                     & 4.06                      & 1.95                     & 1.73                     & \multicolumn{1}{r|}{1.83}  & 3.98                     & 3.61                     & 3.79                      & 1.80                     & 1.67                     & \multicolumn{1}{r|}{1.73}  & 2.85                                    \\
                                                                           &                         &                        & \multicolumn{1}{c|}{RA} & 77.54                    & 45.72                    & 57.52                     & 49.98                    & 29.20                    & \multicolumn{1}{r|}{36.86} & 76.98                    & 44.75                    & 56.60                     & 50.00                    & 29.16                    & \multicolumn{1}{r|}{36.84} & 46.95                                   \\ \cmidrule(l){2-17} 
                                                                           & \multirow{4}{*}{multi}  & \multirow{2}{*}{en}    & \multicolumn{1}{c|}{DP} & 19.53                    & 22.00                    & 20.69                     & 12.53                    & 14.12                    & \multicolumn{1}{r|}{13.28} & 16.41                    & 19.64                    & 17.88                     & 10.51                    & 12.35                    & \multicolumn{1}{r|}{11.35} & 15.80                                   \\
                                                                           &                         &                        & \multicolumn{1}{c|}{RA} & 67.21                    & 64.36                    & 65.75                     & 40.16                    & 38.04                    & \multicolumn{1}{r|}{39.07} & 66.92                    & 58.74                    & 62.56                     & 40.23                    & 35.31                    & \multicolumn{1}{r|}{37.61} & 51.25                                   \\
                                                                           &                         & \multirow{2}{*}{zh}    & \multicolumn{1}{c|}{DP} & 13.31                    & 12.33                    & 12.80                     & 6.86                     & 6.49                     & \multicolumn{1}{r|}{6.67}  & 12.27                    & 12.01                    & 12.13                     & 6.33                     & 6.20                     & \multicolumn{1}{r|}{6.26}  & 9.47                                    \\
                                                                           &                         &                        & \multicolumn{1}{c|}{RA} & 82.38                    & 68.53                    & 74.82                     & 47.75                    & 38.75                    & \multicolumn{1}{r|}{42.78} & 80.78                    & 64.30                    & 71.60                     & 49.60                    & 38.31                    & \multicolumn{1}{r|}{43.23} & 58.11                                   \\ \midrule
\multirow{8}{*}{Kimi}                                                      & \multirow{4}{*}{single} & \multirow{2}{*}{en}    & \multicolumn{1}{c|}{DP} & 24.62                    & 30.16                    & 27.11                     & 13.05                    & 15.65                    & \multicolumn{1}{r|}{14.23} & 19.51                    & 27.57                    & 22.85                     & 9.92                     & 13.21                    & \multicolumn{1}{r|}{11.33} & 18.88                                   \\
                                                                           &                         &                        & \multicolumn{1}{c|}{RA} & 63.26                    & 62.36                    & 62.81                     & 30.85                    & 31.43                    & \multicolumn{1}{r|}{31.14} & 59.41                    & 58.49                    & 58.94                     & 28.53                    & 29.07                    & \multicolumn{1}{r|}{28.79} & 45.42                                   \\
                                                                           &                         & \multirow{2}{*}{zh}    & \multicolumn{1}{c|}{DP} & 29.54                    & 32.02                    & 30.73                     & 15.70                    & 17.22                    & \multicolumn{1}{r|}{16.43} & 28.31                    & 31.59                    & 29.86                     & 14.59                    & 16.35                    & \multicolumn{1}{r|}{15.42} & 23.11                                   \\
                                                                           &                         &                        & \multicolumn{1}{c|}{RA} & 46.88                    & 43.39                    & 45.07                     & 25.98                    & 25.13                    & \multicolumn{1}{r|}{25.55} & 46.07                    & 42.72                    & 44.33                     & 25.40                    & 24.42                    & \multicolumn{1}{r|}{24.90} & 34.96                                   \\ \cmidrule(l){2-17} 
                                                                           & \multirow{4}{*}{multi}  & \multirow{2}{*}{en}    & \multicolumn{1}{c|}{DP} & 37.73                    & 51.20                    & 43.45                     & 21.44                    & 29.85                    & \multicolumn{1}{r|}{24.96} & 30.42                    & 46.25                    & 36.70                     & 15.77                    & 24.77                    & \multicolumn{1}{r|}{19.27} & 31.09                                   \\
                                                                           &                         &                        & \multicolumn{1}{c|}{RA} & 60.13                    & 70.01                    & 64.70                     & 35.08                    & 42.29                    & \multicolumn{1}{r|}{38.35} & 56.07                    & 64.78                    & 60.11                     & 32.00                    & 38.05                    & \multicolumn{1}{r|}{34.76} & 49.48                                   \\
                                                                           &                         & \multirow{2}{*}{zh}    & \multicolumn{1}{c|}{DP} & 67.12                    & 75.76                    & 71.18                     & 31.32                    & 36.54                    & \multicolumn{1}{r|}{33.73} & 65.79                    & 73.18                    & 69.29                     & 29.79                    & 33.80                    & \multicolumn{1}{r|}{31.67} & 51.47                                   \\
                                                                           &                         &                        & RA                      & 70.82                    & 83.01                    & 76.43                     & 37.84                    & 45.33                    & 41.25                      & 67.46                    & 80.01                    & 73.20                     & 35.37                    & 42.11                    & 38.45                      & 57.33                                   \\ \bottomrule
\end{tabular}
}
\caption{Experiment results of closed-source LLMs under single- and multi-document settings in English and Chinese. DP and RA denote direct prompting tracing and retrieval-augmented tracing.}
\label{tab:experiment-appendix2}
\end{table*}

\section{Dataset}
\label{appendix-data}
\autoref{tab:datasets_appendix} presents a detailed statistical overview of our dataset, categorized across multiple dimensions: document type (single vs. multi-document), language (English vs. Chinese), scenario, domain, origin dataset, and average document length.

Our dataset consists of English and Chinese sources, covering multiple scenarios such as news summarization, academic summarization, and question answering. It includes domains such as news, government reports, scientific papers, meetings, and dialogues, ensuring broad coverage across different textual data types. The origin datasets include well-established resources, i.e., Long-Bench, LooGLE, and CRUD.

To account for variations in document length, we report \#AvgLen, which measures the average length of source documents in words for English texts and characters for Chinese texts. Multi-document datasets (e.g., MultiNews) tend to have longer text sequences, while single-document datasets vary significantly based on their domain (e.g., academic papers in LongSum have much longer texts than news articles in EventSum).

\section{Detail Experiment Results}

We present experimental results for open-source and closed-source LLMs under single- and multi-document settings in English and Chinese. \autoref{tab:experiment-appendix1} shows the results for open-source models. \autoref{tab:experiment-appendix2} provides results for closed-source models.

We report metrics for direct prompting (DP) and retrieval-augmented (RA) tracing. Each table includes macro- and micro-averaged precision, recall, and F1 metrics for source tracing (T), relationship classification (R), and an overall F1 score. 

In single-document English tasks,  among open-source models, Qwen2.5-7B with retrieval-augmented tracing achieves the highest F1 (35.02), outperforming other open-source alternatives (e.g., Qwen2.5-14B with 34.50). However, the closed-source Gemini-1.5-Pro obtains an even higher F1 of 48.39 with retrieval, making it the top performer overall in this single-document English scenario. Notably, GPT-4o is quite capable under direct prompting (33.97), exceeding the retrieval-augmented baselines of most open-source LLMs. However, almost all models (open or closed) show significant gains when retrieval is introduced. 

In single-document Chinese tasks, among open-source models, DeepSeek-V3 (RA) leads with an F1 score of 44.52, outperforming Qwen and Llama. Among closed-source models, GPT-4o (DP) scores 42.99, while Gemini-1.5-Pro (RA) gets higher at 46.95. Although these closed-source models exceed most open-source options except DeepSeek-V3, GPT-4o also performs well without retrieval, scoring 43.61 with direct prompting, even better than its RA variant. In contrast, Gemini relies heavily on retrieval, as shown by its sharp jump from a very low direct prompting score of 2.85 to 46.95 when retrieval is applied. This highlights the varying levels of dependence on reducing candidates among different models.

In multi-document English tasks, Qwen2.5-14b (RA) leads among open-source models with an F1 score of 44.54, slightly ahead of DeepSeek-V3 (43.39). However, the closed-source Gemini-1.5-Pro gets the top score with 51.25, outperforming GPT-4o (48.34) and Kimi (49.48). GPT-4o also shows strong performance without retrieval, scoring 41.26, while Gemini struggles with a much lower 15.80. This suggests that GPT-4o is naturally better at direct prompting, whereas Gemini and Kimi depend more on retrieved context to handle complex multi-document provenance.

In multi-document Chinese tasks, DeepSeek-V3 (RA) leads open-source models with an F1 score of 54.52, far ahead of Qwen2.5-14B (47.51). However, GPT-4o achieves the best overall with 61.09, just ahead of Gemini-1.5-Pro (58.11) and Kimi (57.33). This highlights GPT-4o's strong ability to handle multi-source Chinese text. 

In all, in single-document tasks, Qwen2.5-7B and DeepSeek-V3 emerge as strong open-source choices for English and Chinese, respectively, yet Gemini-1.5-Pro can outperform them once retrieval is incorporated. GPT-4o stands out for its relatively high direct-prompting scores across both languages, showing strong built-in tracing capabilities. Under multi-document conditions, the complexity increases, and the top results often come from closed-source solutions (e.g., Gemini-1.5-pro, GPT-4o, Kimi), although Qwen2.5-14b and DeepSeek-V3 hold their own in the open-source domain. Models integrating retrieval, whether open- or closed-source, generally exhibit greater gains and more accurate sentence-level provenance.

\subsection{Confusion Matrix Analysis Details}

The observed discrepancies in total counts between confusion matrices for different methods are attributed to the following methodological factors: (1) The retrieval-augmented method may fail to retrieve certain sentences from the source documents, leading to variations in the number of ground-truth sentences available for classification. (2) The retrieval-augmented and direct prompting methods trace different sets of source sentences due to their distinct retrieval mechanisms. Sentences that remain untraced by either method are excluded from the subsequent relationship classification task, resulting in different sample sizes across experimental conditions. It is worth noting that we employ the pass@5 evaluation metric for all experimental assessments to ensure consistent and robust performance measurement.

\section{Prompts}

To prevent large language models from mislabeling, the pre-labeling process of GPT-4o adopts a more fine-grained classification, specifically as: Copy, Paraphrase, Summary, Inference, Expansion, Fusion, Distillation, Reordering, Negation, Generalization. And the mapping between pre-labels and final labels is shown in \autoref{tab:label_mapping}.

\begin{figure}[h]
\setlength{\abovecaptionskip}{0.12cm}    
\setlength{\belowcaptionskip}{-0.6cm}
  \centering 
  \includegraphics[width=0.4 \textwidth]{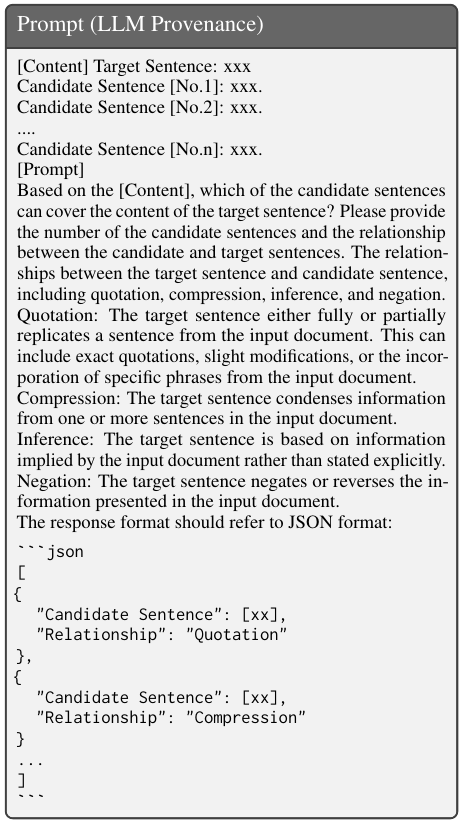}\\
  \caption{Prompt for LLM provenance in experiments. \label{fig:prompt}}
\end{figure}

\begin{figure*}[t]
\setlength{\abovecaptionskip}{0.12cm}    
\setlength{\belowcaptionskip}{-0.6cm}
  \centering 
  \includegraphics[width=0.96 \textwidth]{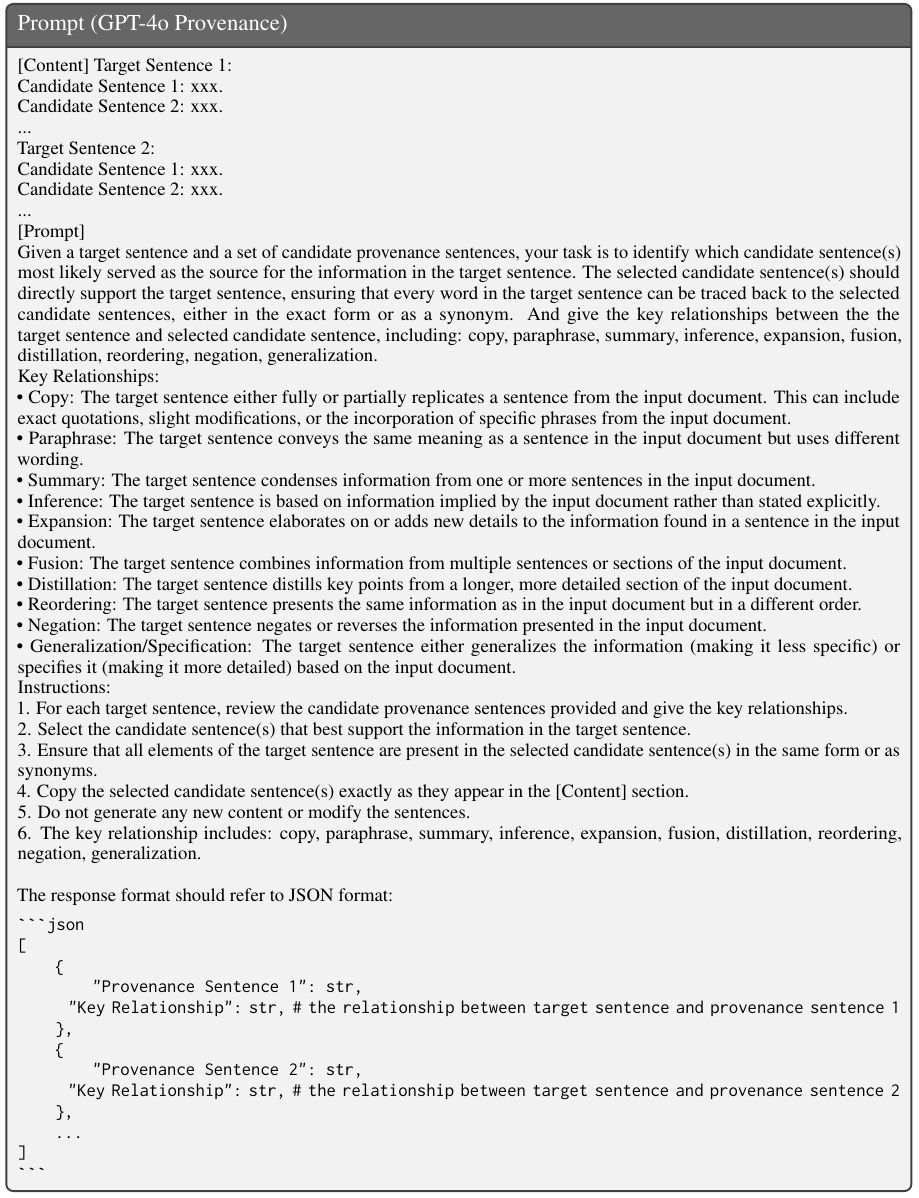}\\
  \caption{Prompt for GPT4o provenance in annotation. \label{fig:prompt2}}
\end{figure*}

\begin{figure*}[t]
\setlength{\abovecaptionskip}{0.12cm}    
\setlength{\belowcaptionskip}{-0.6cm}
  \centering 
  \includegraphics[width=0.98 \textwidth]{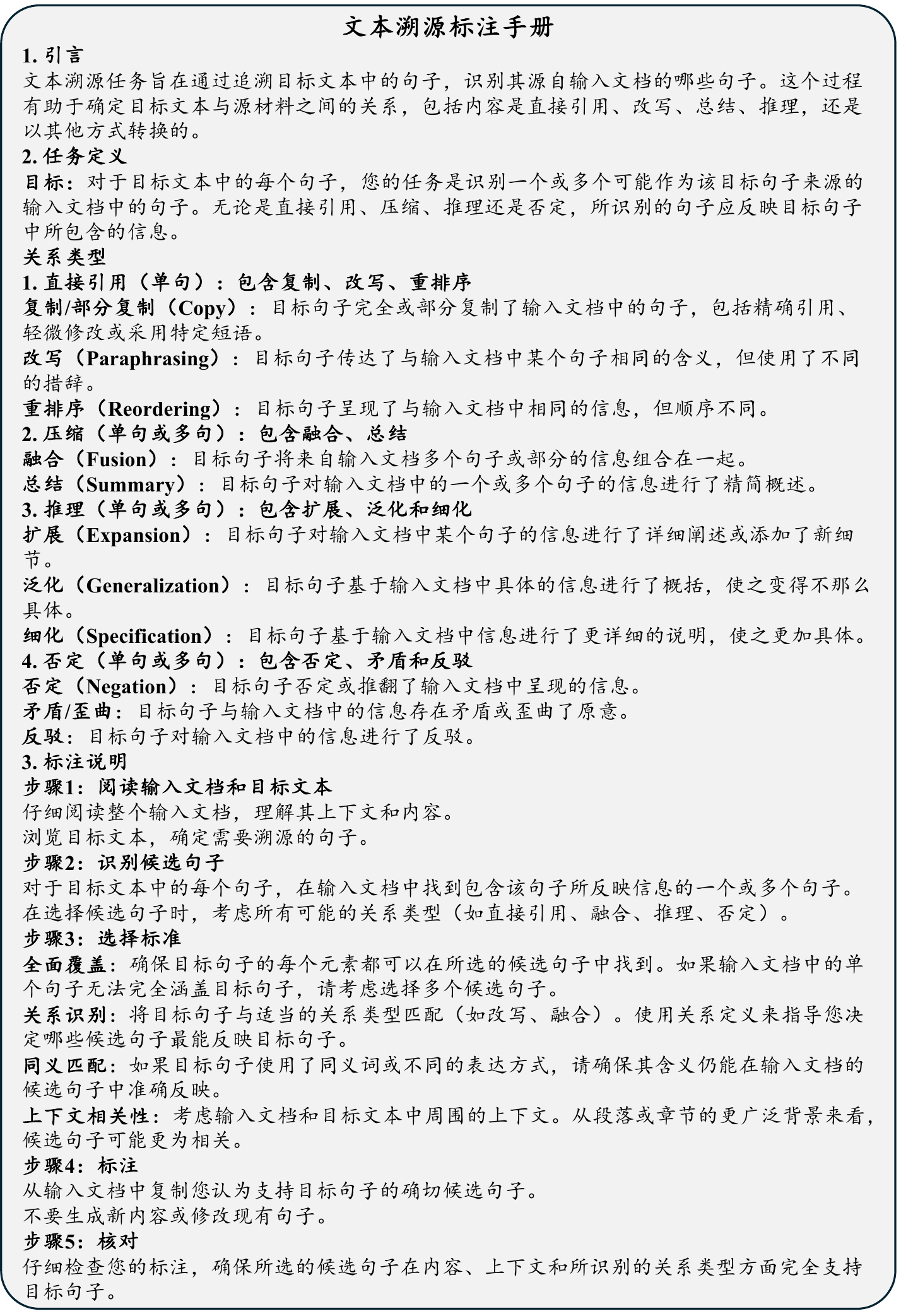}\\
  \caption{The guideline for human annotation. \label{fig:guideline}}
\end{figure*}

\end{document}